\documentclass[conference]{IEEEtran}
\IEEEoverridecommandlockouts
\usepackage{cite}
\usepackage{amsmath,amssymb,amsfonts}
\usepackage{algorithmic}
\usepackage{graphicx}
\usepackage{textcomp}
\usepackage{xcolor}
\usepackage{tikz}
\usepackage{caption}
\usepackage{float}
\usepackage{subcaption}
\usepackage{placeins}
\usepackage{tikz}
\usepackage{booktabs} 
\usetikzlibrary{shapes.geometric, arrows, positioning}
\usepackage[colorlinks=true, linkcolor=blue, urlcolor=blue, citecolor=blue]{hyperref}
\def\BibTeX{{\rm B\kern-.05em{\sc i\kern-.025em b}\kern-.08em
    T\kern-.1667em\lower.7ex\hbox{E}\kern-.125emX}}
\begin{document}

\title{Fine-Tuning Stable Diffusion XL for Stylistic Icon Generation: A Comparison of Caption Size}

\author{
\IEEEauthorblockN{Youssef Sultan\IEEEauthorrefmark{1}, Jiangqin Ma\IEEEauthorrefmark{1}, Yu-Ying Liao\IEEEauthorrefmark{1}}
\IEEEauthorblockA{
\textit{College of Engineering}, \\
\textit{Georgia Institute of Technology}, \\
801 Atlantic Dr NW, Atlanta, 30332, GA, USA \\
\{ysultan,  jma416, yliao1219\}@gatech.edu
 \IEEEauthorblockA{\IEEEauthorrefmark{1} These authors contributed equally to this work.}
}
}

\maketitle

\begin{abstract}
In this paper, we show different fine-tuning methods for Stable Diffusion XL; this includes inference steps, and caption customization for each image to align with generating images in the style of a commercial 2D icon training set. We also show how important it is to properly define what "high-quality" really is especially for a commercial-use environment. As generative AI models continue to gain widespread acceptance and usage, there emerge many different ways to optimize and evaluate them for various applications \cite{b1}. Specifically text-to-image models, such as Stable Diffusion XL \cite{b2} and DALL-E 3 \cite{b3} require distinct evaluation practices to effectively generate high-quality icons according to a specific style. Although some images that are generated based on a certain style may have a lower FID score (better), we show how this is not absolute in and of itself even for rasterized icons. While FID \cite{b5}\cite{b6} scores reflect the similarity of generated images to the overall training set, CLIP \cite{b4} scores measure the alignment between generated images and their textual descriptions. We show how FID scores miss significant aspects, such as the minority of pixel differences that matter most in an icon, while CLIP scores result in misjudging the quality of icons. The CLIP model's understanding of "similarity" is shaped by its own training data; which does not account for feature variation in our style of choice. Our findings highlight the need for specialized evaluation metrics and fine-tuning approaches when generating high-quality commercial icons, potentially leading to more effective and tailored applications of text-to-image models in professional design contexts.
\end{abstract}


\section{Introduction}
Diffusion models have shifted the field of machine learning by enabling the creation of supposed high-fidelity synthetic data. They work by iteratively denoising data through a stochastic process, then transforming that random noise into coherent and realistic outputs \cite{b7}. Denoising Diffusion Probabilistic Models (DDPMs) model the data distribution with precision through a forward and reverse process. Within the forward process, Gaussian noise is added to the data, mapping it to a noise distribution. However, the reverse process learns how to denoise this distribution. The results are a reconstruction of the original data distribution, but in a different form. This different form outlines the generative aspect of the model, allowing it to capture intricate details and produce something different yet coherent and in some cases realistic.

The main caveat with these models unfortunately is the efficiency, training time and resources needed in order to train or fine-tune them tailored to a specific use-case. This has led to additional works further expanding into the field of efficiency and efficacy where techniques like weight quantization in "BitsFusion" \cite{b8} have significantly reduced these requirements while maintaining performance. Advancements of diffusion models for tasks tailored to commercial or user-driven use-cases have been a critical area of research. "DreamBooth" \cite{b9} and "StyleDrop" \cite{b11} modify text-to-image diffusion models to generate content that is aligned with a specific style or aesthetic, highlighting an easier route for such use-cases. 

Additionally, high-resolution image synthesis as discussed by Rombach et al. \cite{b10} provides a great vantage point of latent diffusion models (LDMs) in producing detailed and high-quality images. LDMs work in a compressed latent space as opposed to directly in pixel space, which is better for efficiency, model training and inference (image-generation). Despite such compression, they discard noise and redundant information within the image, and capture essential data characteristics through variational autoencoders \cite{b13}. 

The works of Meng et al. in "SDEdit" \cite{b12} also introduce guided image synthesis and editing using stochastic differential equations, adding a new barometer to the playing field. By manipulating the trajectory of the image generation process through user-defined synthesis criteria, it allows for a balance between deterministic control and stochastic control at inference time. This of course translates to the tailoring of desired image characteristics whether it be specific styles, textures, color schemes or structural elements of the image.

Stable Diffusion XL \cite{b2}, also a latent diffusion model, leverages more parameters in its architecture and offers improved fine-tuning capabilities. The key advancement in Stable Diffusion XL is its refinement model, which is employed at the end to enhance the visual fidelity of the generated image. Its ability to adhere to specific visual styles make it particularly well-suited for tasks like stylistic icon generation. This characteristic, combined with its advanced architecture, makes Stable Diffusion XL an ideal candidate for our study on generating high-quality commercial icons.

\section{Methods}
In this section we detail the approaches and techniques used to fine-tune Stable Diffusion XL in order to generate icons pertaining to a specific preset style for commercial use. We explore various aspects including inference steps, and the integration of different sized captions to align generated images with our specific icon style. Additionally, we compare the outcomes using the same captions with DALL-E 3 to understand the CLIP and FID score comparison which also shows how these metrics fluctuate relatively to the actual quality of the icons assessed by a human.
\begin{figure}[h!]
  \centering
  \includegraphics[width=\linewidth]{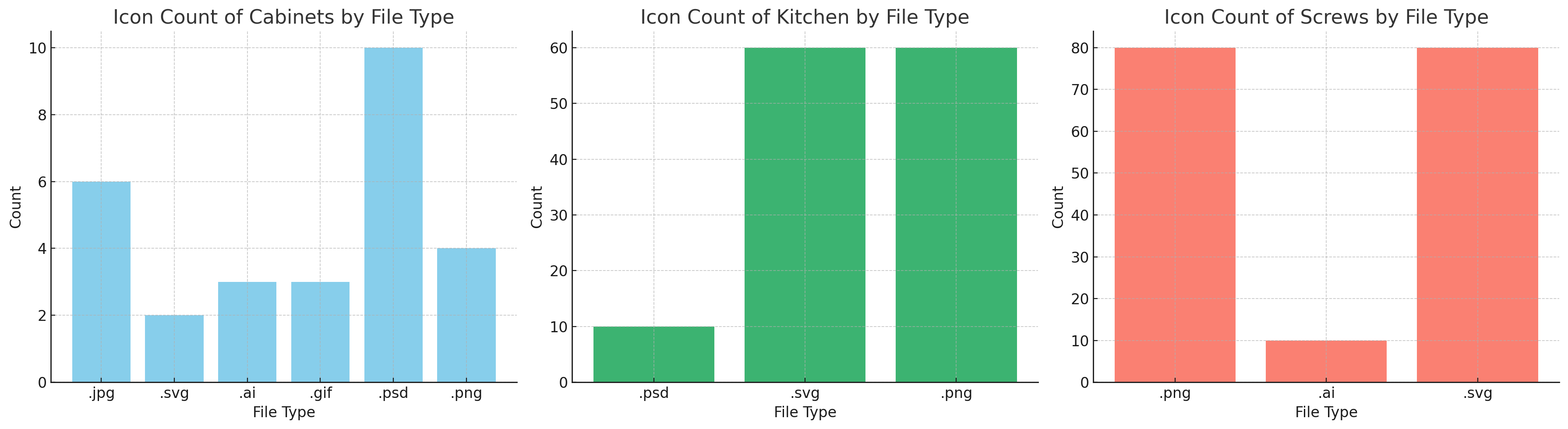}
  \caption{Distribution of file types for icon images related to kitchen cabinets, kitchens, and screws in the provided commercial data.}
  \label{fig:icon_distribution}
\end{figure}

We utilized a dataset from The Home Depot containing images of various icon types categorized by their descriptions within the file name. For example one image icon may be a "phillips head screw" with an icon of a 2D phillips head screw. The focus here was to distinguish screws and kitchen cabinets for our experiments, so we cleaned the data in order to filter out any unrelated images to those two categories. This left us with 42 icon images related to screws and 17 icon images related to kitchen cabinets. The overall distribution of the file types regarding these images can be seen in Figure~\ref{fig:icon_distribution}.

\begin{figure}[htbp]
\centerline{\includegraphics[width=0.55\linewidth]{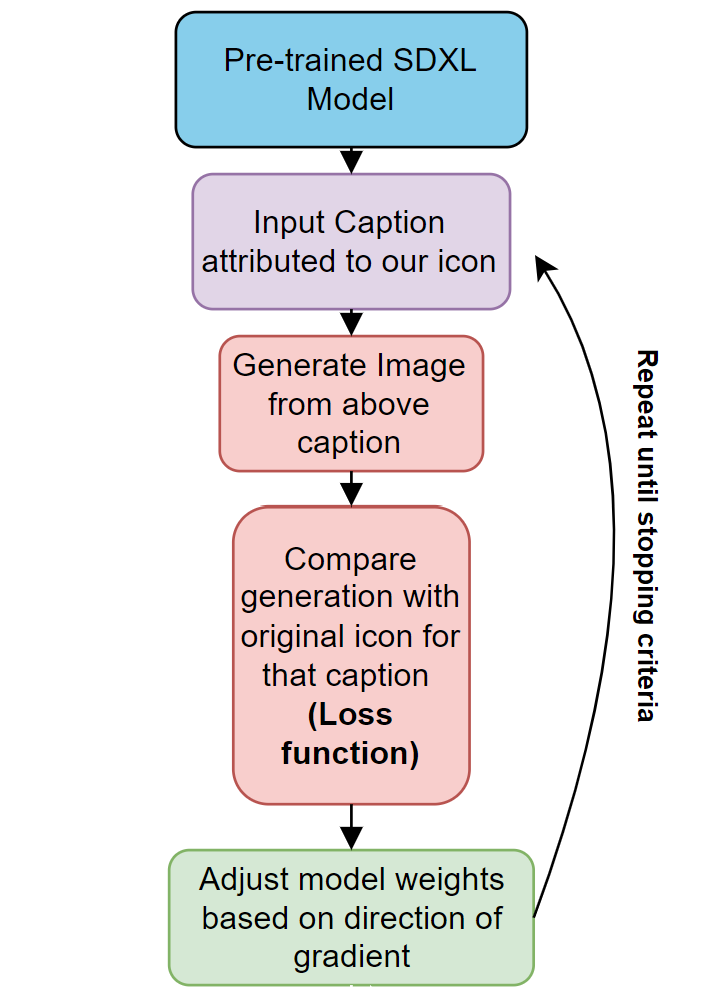}}
\caption{A visual representation of how the model learns to generate images according to a style}
\label{inferencepipeline}
\end{figure}

Training these types of models requires an image paired with a caption. In the phillips head screw example, our goal is to have a caption that best describes what this image represents. Therefore, for every image, we have manually defined a custom prompt for that image in a metadata.jsonl file. The purpose of this is to fine-tune the model, where "fine-tuning" is essentially an adjustment of the model weights to get more of a match between the generated image and our original image (icon in the training set). From a high-level overview, the sequence of the fine-tuning can be interpreted in Figure~\ref{inferencepipeline}.

\begin{table}[h!]
\centering
\caption{Training Parameters for the Models}
\begin{tabular}{@{}ll@{}}
\texttt{mixed\_precision} & fp16 \\
\texttt{resolution} & 1024 \\
\texttt{train\_batch\_size} & 1 \\
\texttt{gradient\_accumulation\_steps} & 3 \\
\texttt{gradient\_checkpointing} & True \\
\texttt{learning\_rate} & 1e-4 \\
\texttt{snr\_gamma} & 5.0 \\
\texttt{lr\_scheduler} & constant \\
\texttt{lr\_warmup\_steps} & 0 \\
\texttt{use\_8bit\_adam} & True \\
\texttt{max\_train\_steps} & 500 \\
\texttt{checkpointing\_steps} & 717 \\
\texttt{seed} & 0 \\ \bottomrule
\end{tabular}
\label{tab:training_params}
\end{table}

For screw icons, we trained three versions of Stable Diffusion XL using the same set of images but with different prompts to see the variation in icon quality, FID score and CLIP score. The model variants were short prompts with class images, short prompts and long prompts. The short prompt with class images leveraged short keywords and 20 high-quality Home Depot head-type icon images. The short prompt contained short keyword prompts whereas the long prompt contained detailed descriptions of the training dataset images. A fourth model used for comparison purposes was the DALL-E 3, which was accessed via API to generate images using the same inference prompt. The purpose of including DALL-E 3 was to understand performance metric variation relative to the quality by human assessment. The parameters used for the three Stable Diffusion XL models can be seen in Table~\ref{tab:training_params}.

Prior-preservation loss is effective in encouraging output diversity and in overcoming language-drift in "DreamBooth" \cite{b9}. Therefore, we add class-specific prior preservation loss to the Stable Diffusion XL model with short prompt during training using class images. This approach requires additional class images and careful balancing between instance-specific and class-generic images. By default, the prior\_loss\_weight is set to 1. This parameter controls the weight of the prior preservation loss during training, which helps to preserve the generalization capabilities of the model by penalizing deviations from the class-specific distribution. Class images are a set of reference images used to guide the training process of a machine learning model. These images are used to help the model understand the style and characteristics of the icons we want to generate. 

Our results show that using the default prior\_loss\_weight did not increase performance significantly, since training with class images has better FID score, but worse CLIP score. Consequently, we needed to experiment with this value to find the optimal setting. This approach requires significant computational resources and time to find the optimal prior\_loss\_weight value and class images. We tried several different class images but were unable to improve performance. Due to limited computational resources and time constraints, we did not continue pursuing this approach.

To reproduce our results, we also conduct this on a public icon dataset including 42 images. The public dataset includes icons downloaded from online resources, primarily using Google search, ChatGPT \cite{chatgpt2023}, and Gemini \cite{gemini2023}. The prompts used for training are also the same as those used for The Home Depot dataset, as we use the same head type and drive style. 

For each model we generate 10 images and use the associated prompts to calculate the average CLIP score. To obtain the FID score differences between each model the same was done for 7 icons with their corresponding generations.

For kitchen cabinet icons, we also trained three versions of Stable Diffusion XL using the same set of images but with different prompts to see the variation in icon quality, FID score and CLIP score. In terms of training data, since there are a limited number of kitchen cabinet icons provided by The Home Depot, we gather multiple free kitchen icons from the internet in a similar manner to our screws methodology which brings our final dataset to be 42 icons. In terms of class images, we leverage 100 icons of appliances and furniture from Home Depot, all with the style we're hoping to generate.

In terms of model variants, as mentioned before there are long prompts with class images, short prompts, and long prompts. Long prompts follow the format of "a photo of TOK kitchen cabinet icon, $<$style$>$, Icon of a $<$cabinet type$>$ kitchen cabinet in $<$color$>$ with $<$detail description of the cabinet type or corner cabinet type$>$", while short prompts follow the same order of long prompts but with keywords, for example "$<$style$>$, $<$cabinet type$>$, $<$color$>$, $<$description $>$". Finally, as a comparison we also leveraged DALL-E 3 to generate images using the same inference prompt for both short and long prompts. Throughout our experiments, we used 'TOK' as a unique identifier for cabinets in our prompts. This identifier acts as a consistent reference point for the model. Additionally, we used the syntax $<$variable$>$ to denote elements in the prompt that we wanted the model to learn and vary. For example, a prompt might include $<$color$>$ or $<$cabinet type$>$, allowing the model to generate diverse cabinet icons by learning to substitute different colors or cabinet types. This approach, inspired by \cite{b9}, enables the creation of novel, photo-realistic images of the subject, contextualized in diverse scenes.

For each model after fine-tuning we generated 40 images using 10 different prompts (inference of 4 for each prompt). The associated prompt used for icon generation was saved in separate txt files. In addition, we selected the best image(s) per prompt and calculated the average CLIP score for each model using the txt files with prompt and corresponding image. This leaves us with 3 icons from our training set to compare with 3 generated images to then calculate the average FID score. Since there were not a lot of kitchen cabinet icons from The Home Depot dataset, we could only calculate the FID on a basis of 3 images.

\section{Results}
\subsection{Models trained on commercial data for screws}
%
%
%
%
%
%
%

\begin{figure}[htbp]
\centerline{\includegraphics[width=1\linewidth]{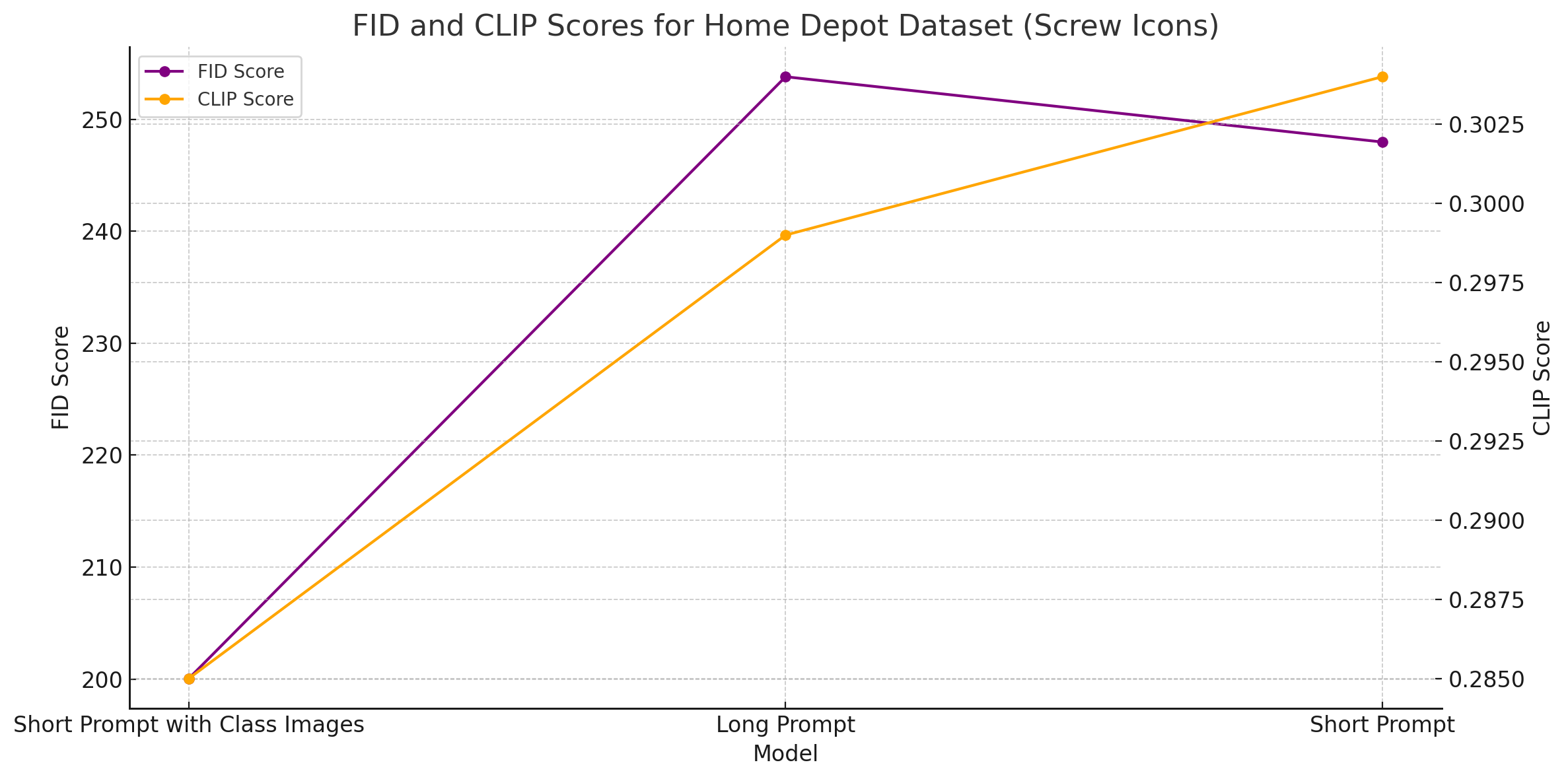}}
\caption{FID and CLIP scores of screw icons generated using commercial icons for training (non-public data)}
\label{screwfidclip}
\end{figure}
Using the model trained on commercial data with longer captions, shorter captions, and class images for screw icon generation, the FID and CLIP scores can be viewed in Figure~\ref{screwfidclip}. Here it can be seen that the short prompt with class images performs the best in terms of CLIP and FID.

\begin{figure}[htbp]
    \centering
    \begin{subfigure}{0.22\textwidth}
        \centering
        \includegraphics[width=\linewidth]{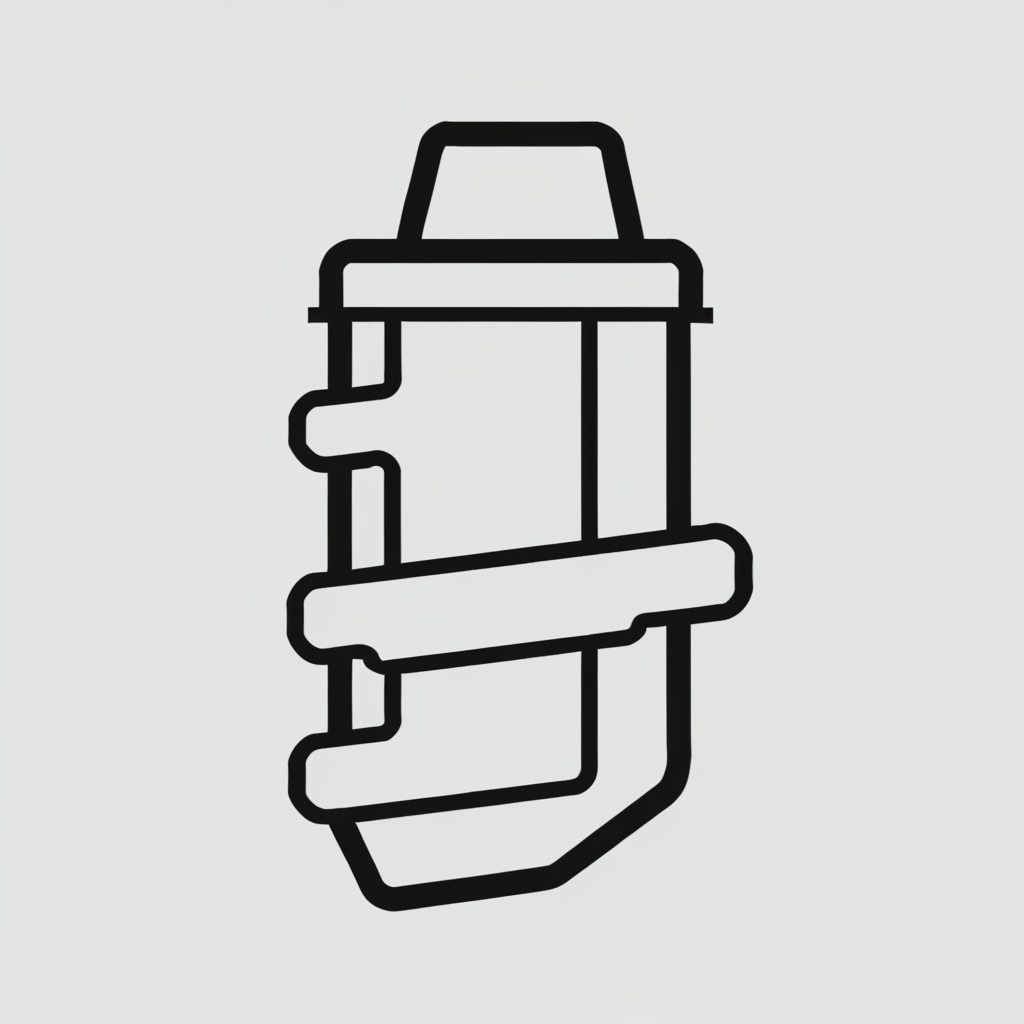}
        \caption{Hex head screw icon}
        \label{fig:hex_head_long}
    \end{subfigure}
    \hfill
    \begin{subfigure}{0.22\textwidth}
        \centering
        \includegraphics[width=\linewidth]{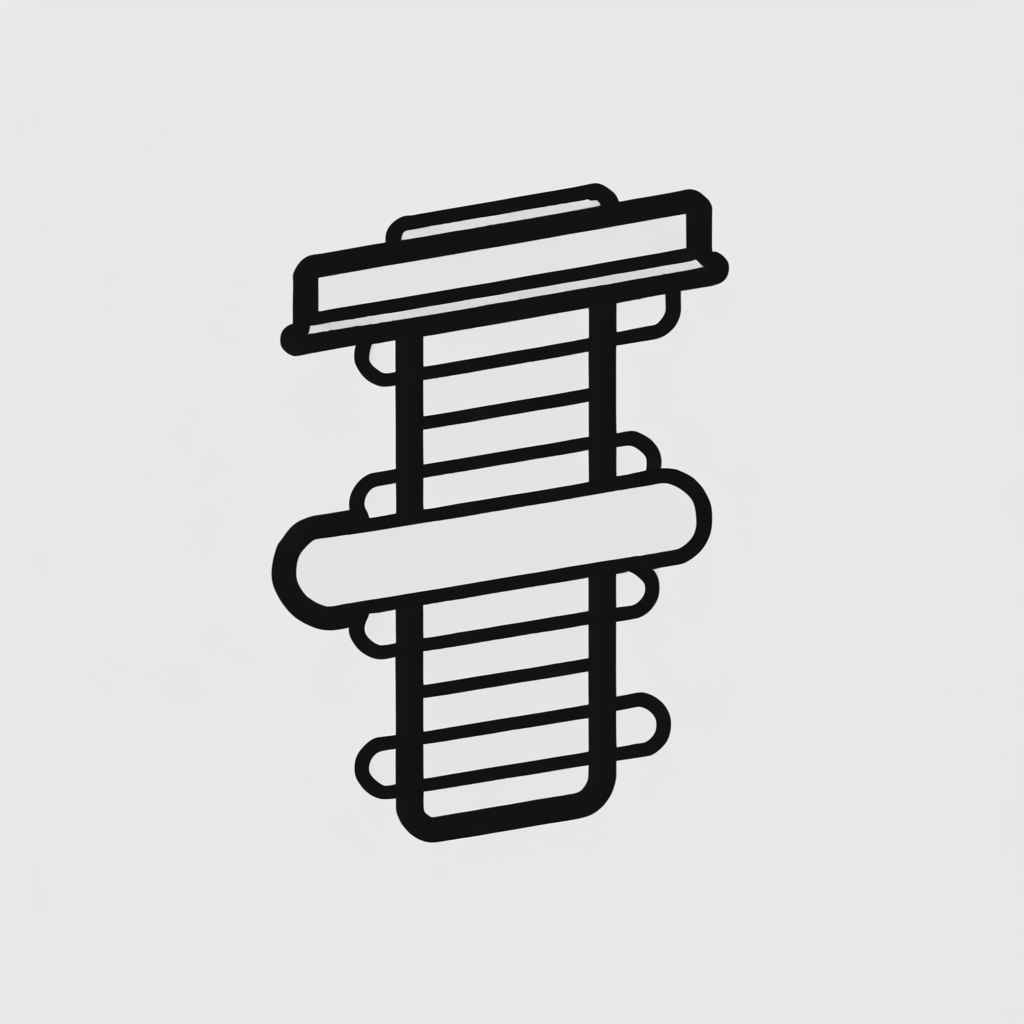}
        \caption{Round head screw icon}
        \label{fig:round_head_long}
    \end{subfigure}
    \hfill
    \begin{subfigure}{0.22\textwidth}
        \centering
        \includegraphics[width=\linewidth]{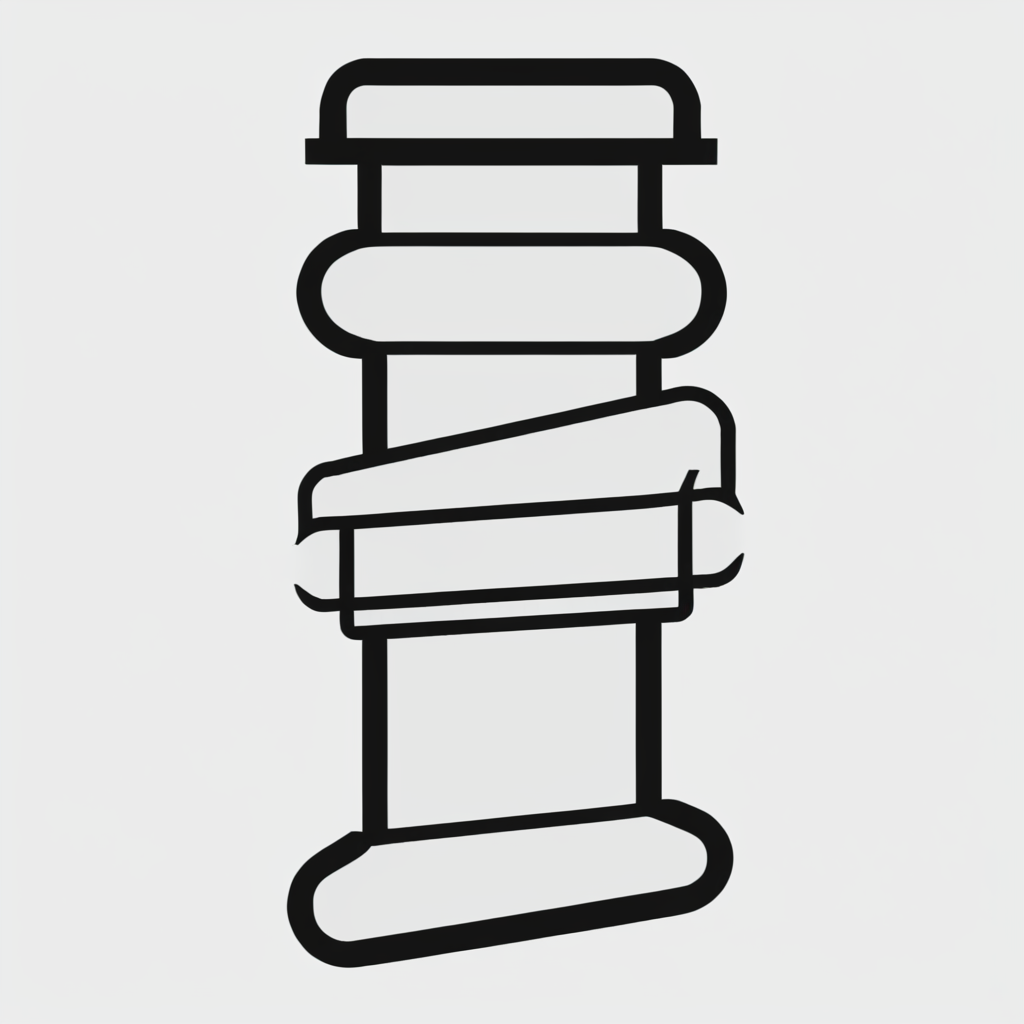}
        \caption{Phillips drive oval head screw icon}
        \label{fig:phillips_oval_long}
    \end{subfigure}
    \hfill
    \begin{subfigure}{0.22\textwidth}
        \centering
        \includegraphics[width=\linewidth]{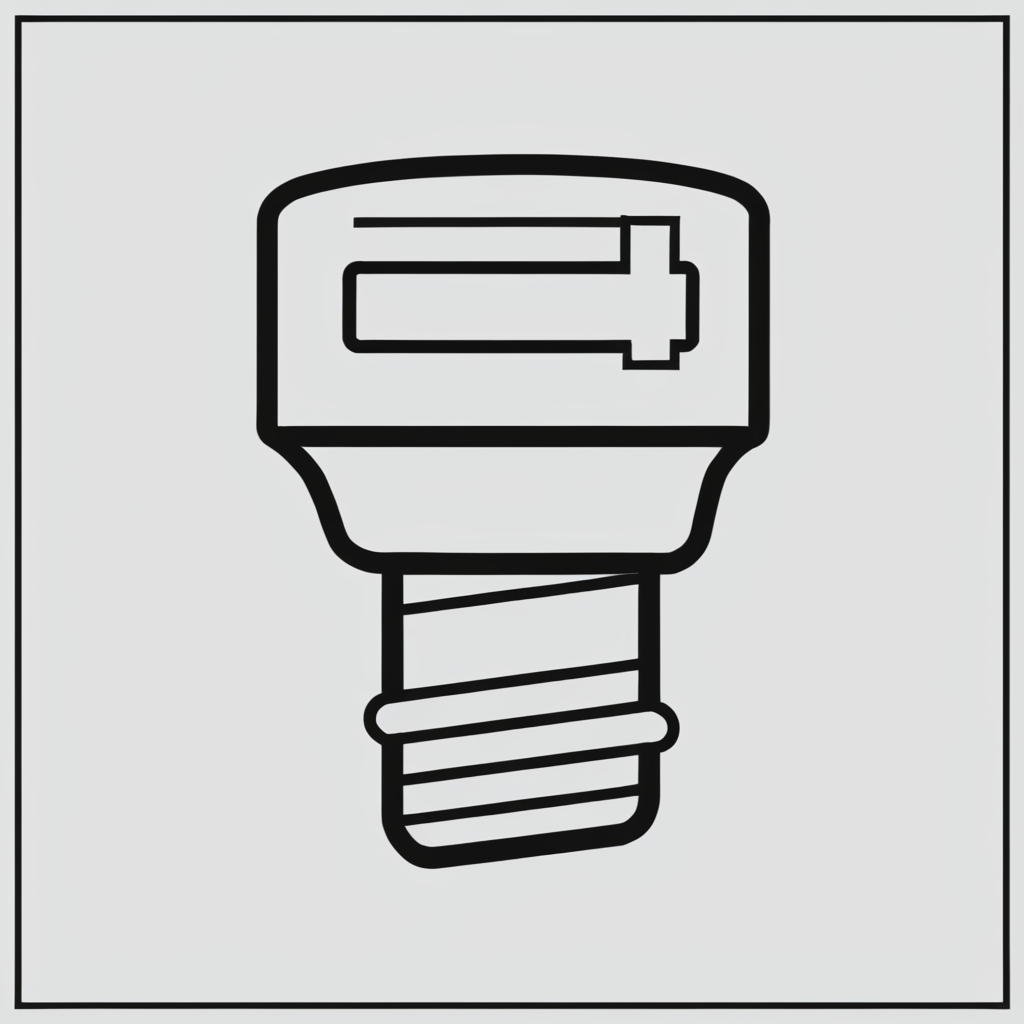}
        \caption{Spanner drive pan head screw icon}
        \label{fig:spanner_pan_long}
    \end{subfigure}
    \caption{Example generated icons based on commercial training + long prompts (each caption above is what was used at inference time)}
    \label{fig:home_depot_long_prompts}
\end{figure}

A sample of 4 generated images along with the inference prompts passed on to the fine-tuned model can be seen in Figure~\ref{fig:home_depot_long_prompts}. As mentioned previously, the FID score is based on an average of 7 images, while the CLIP score is based on an average of 10 images.

\begin{figure}[htbp]
    \centering
    \begin{subfigure}{0.22\textwidth}
        \centering
        \includegraphics[width=\linewidth]{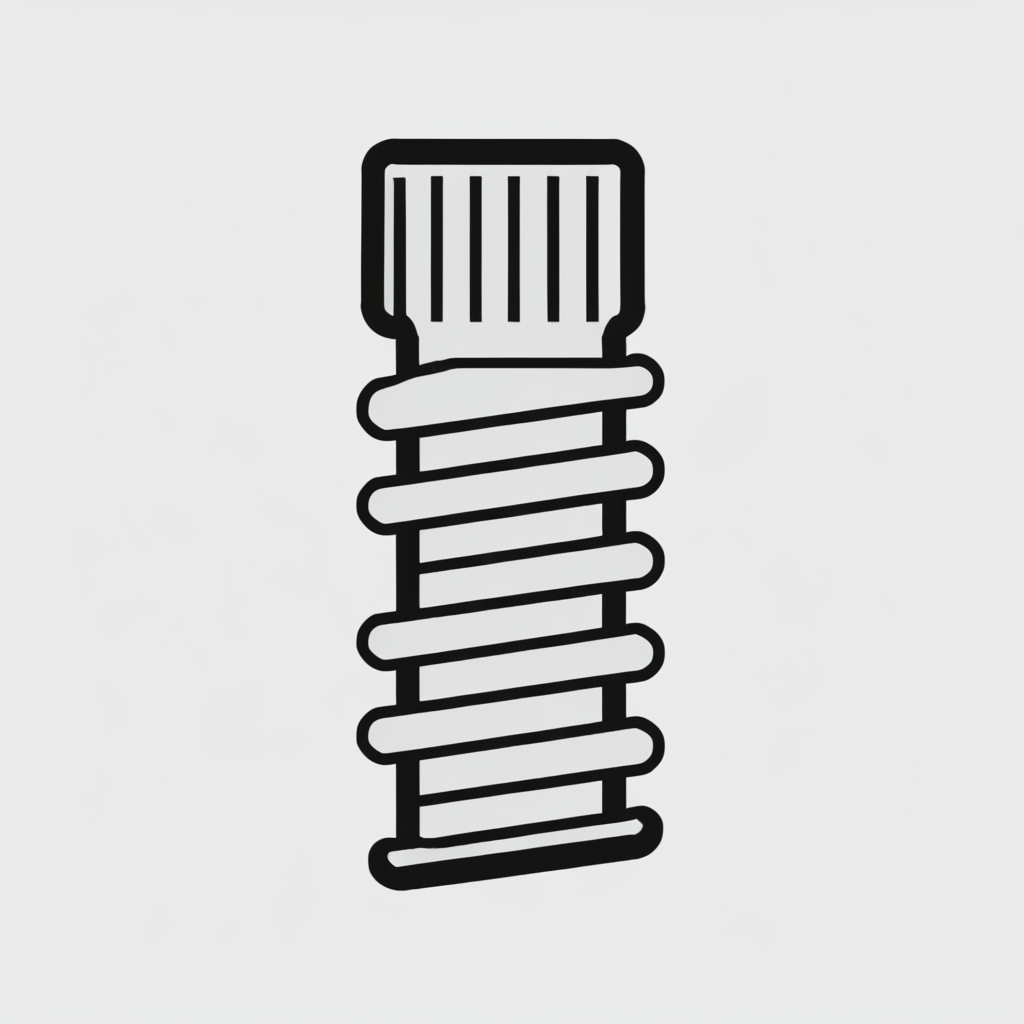}
        \caption{Hex head screw icon}
        \label{fig:hex_head_short}
    \end{subfigure}
    \hfill
    \begin{subfigure}{0.22\textwidth}
        \centering
        \includegraphics[width=\linewidth]{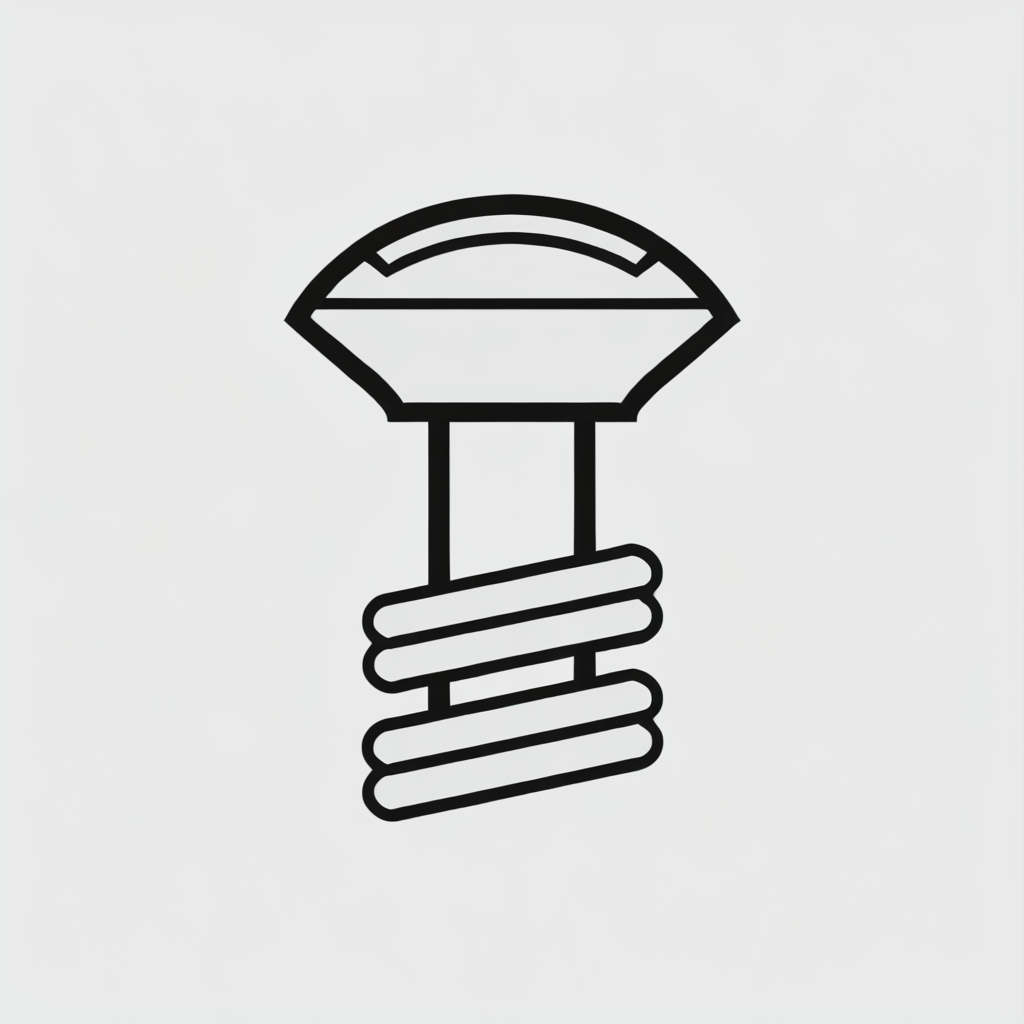}
        \caption{Round head screw icon}
        \label{fig:round_head_short}
    \end{subfigure}
    \hfill
    \begin{subfigure}{0.22\textwidth}
        \centering
        \includegraphics[width=\linewidth]{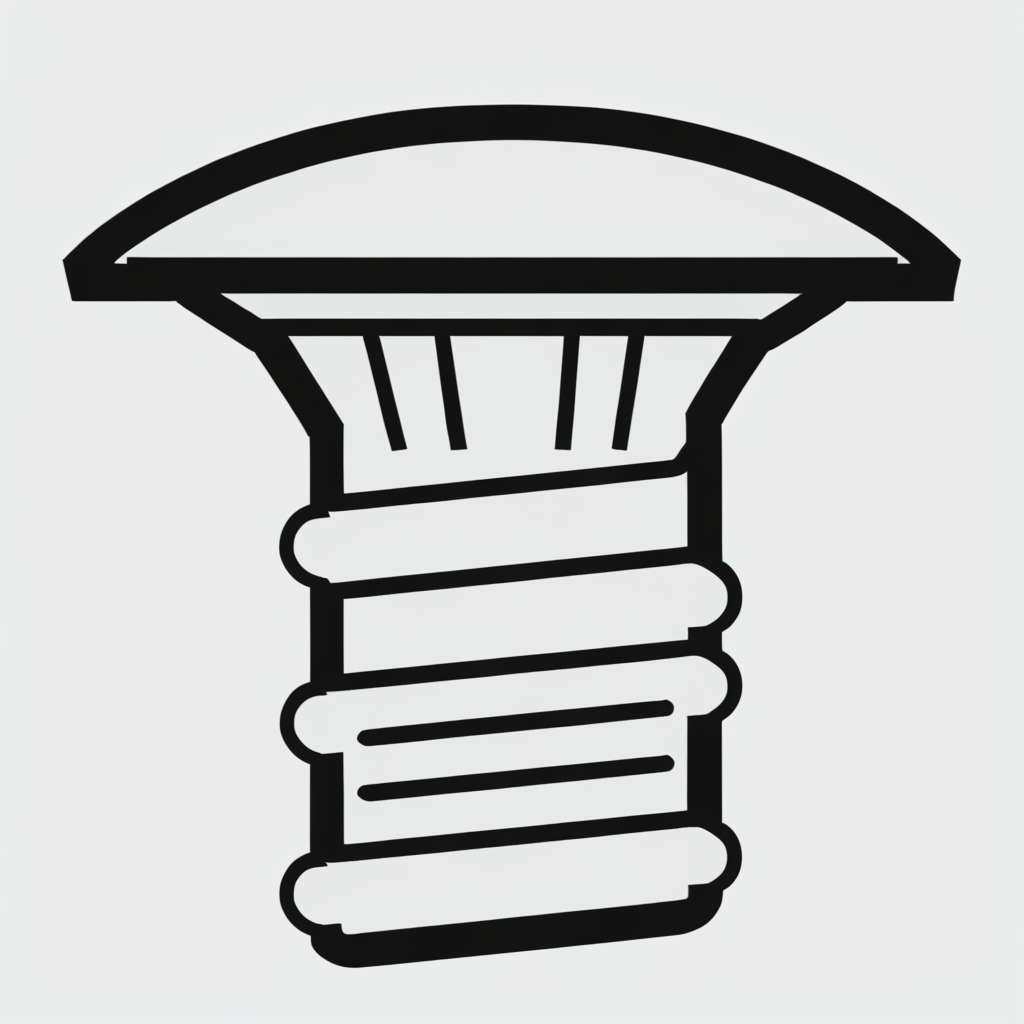}
        \caption{Phillips drive oval head screw icon}
        \label{fig:phillips_oval_short}
    \end{subfigure}
    \hfill
    \begin{subfigure}{0.22\textwidth}
        \centering
        \includegraphics[width=\linewidth]{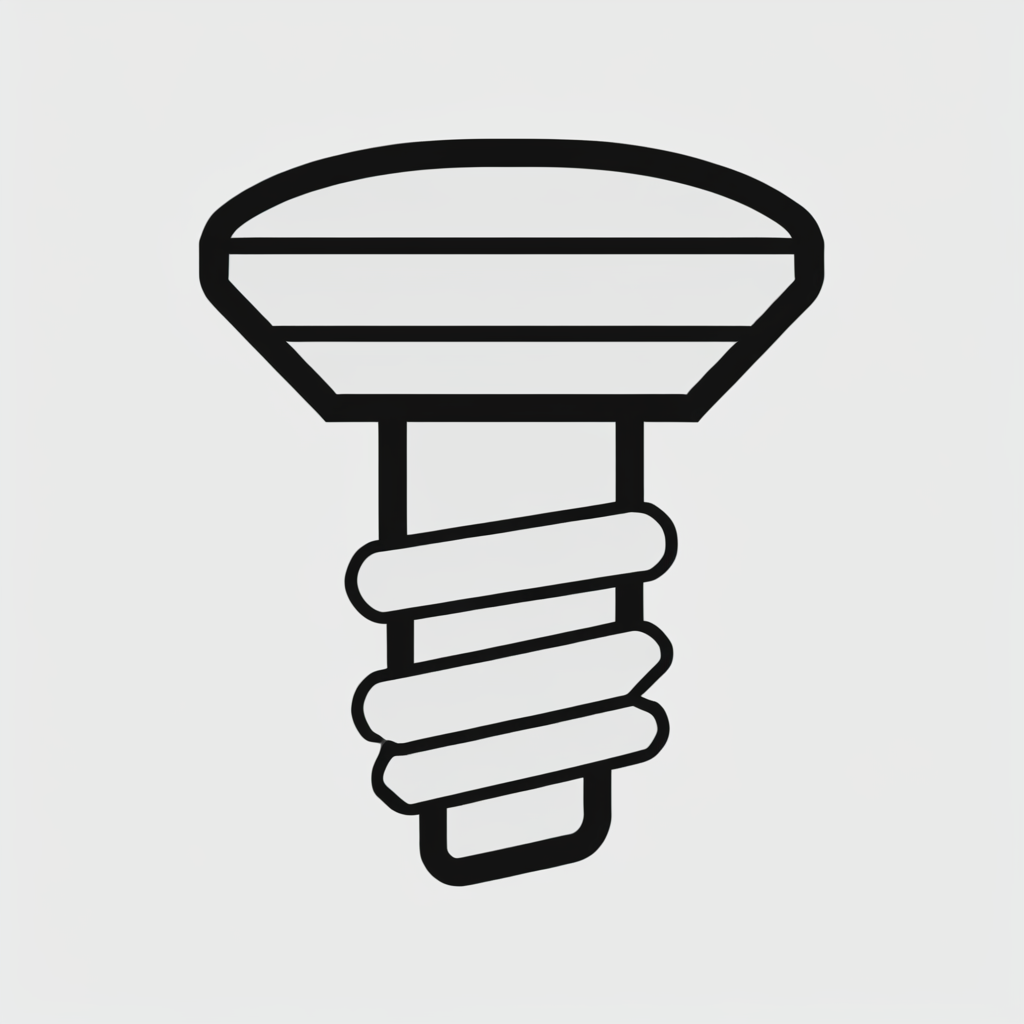}
        \caption{Spanner drive pan head screw icon}
        \label{fig:spanner_pan_short}
    \end{subfigure}
    \caption{Example generated icons based on commercial training + short prompts (each caption above is what was used at inference time)}
    \label{fig:home_depot_short_prompts}
\end{figure}

For the model trained on short prompts, it can be seen that better performance metrics accurately correlate with the human assessment of better screws. That is, the hex screw icon with a shorter prompt in Figure~\ref{fig:home_depot_short_prompts} generated looks more coherent as opposed to the generated image in the long prompt.

\begin{figure}[htbp]
    \centering
    \begin{subfigure}{0.22\textwidth}
        \centering
        \includegraphics[width=\linewidth]{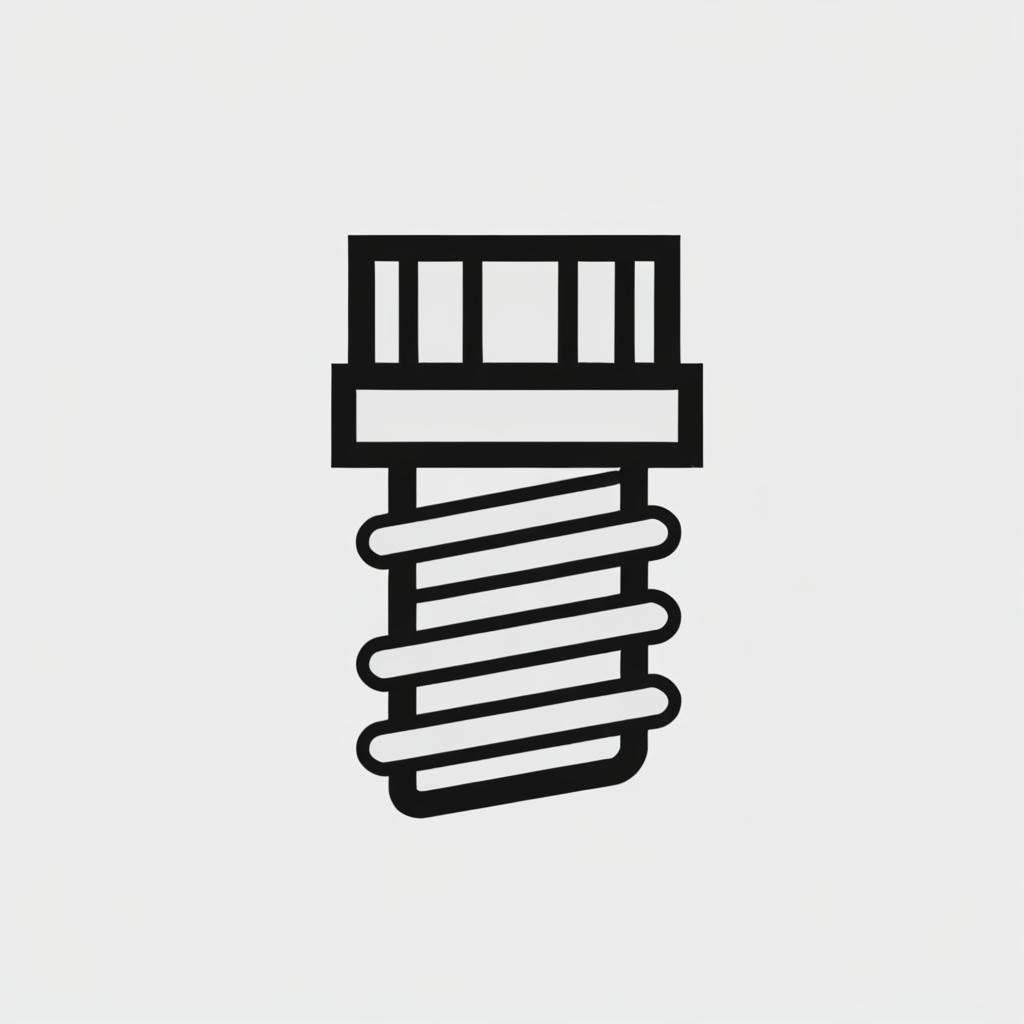}
        \caption{Hex head screw icon}
        \label{fig:class_hex_head_short}
    \end{subfigure}
    \hfill
    \begin{subfigure}{0.22\textwidth}
        \centering
        \includegraphics[width=\linewidth]{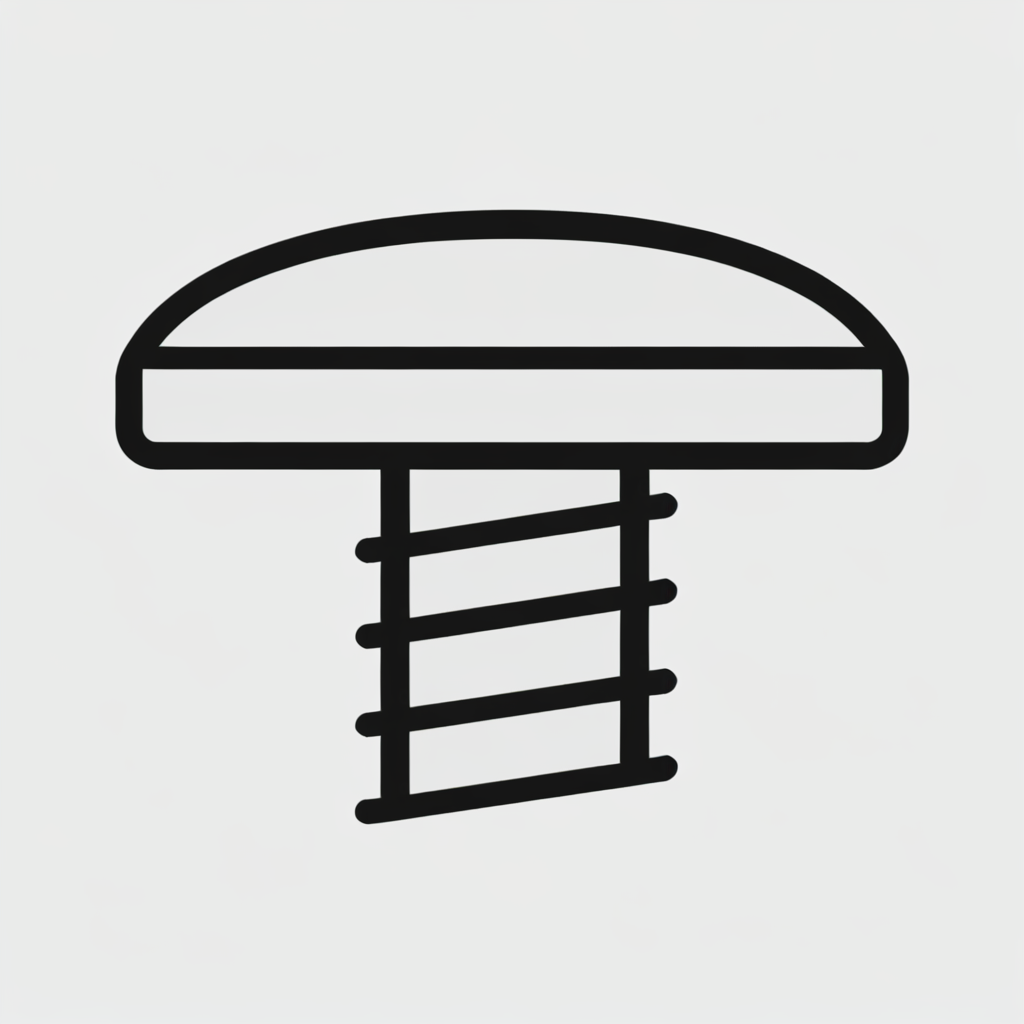}
        \caption{Round head screw icon}
        \label{fig:class_round_head_short}
    \end{subfigure}
    \hfill
    \begin{subfigure}{0.22\textwidth}
        \centering
        \includegraphics[width=\linewidth]{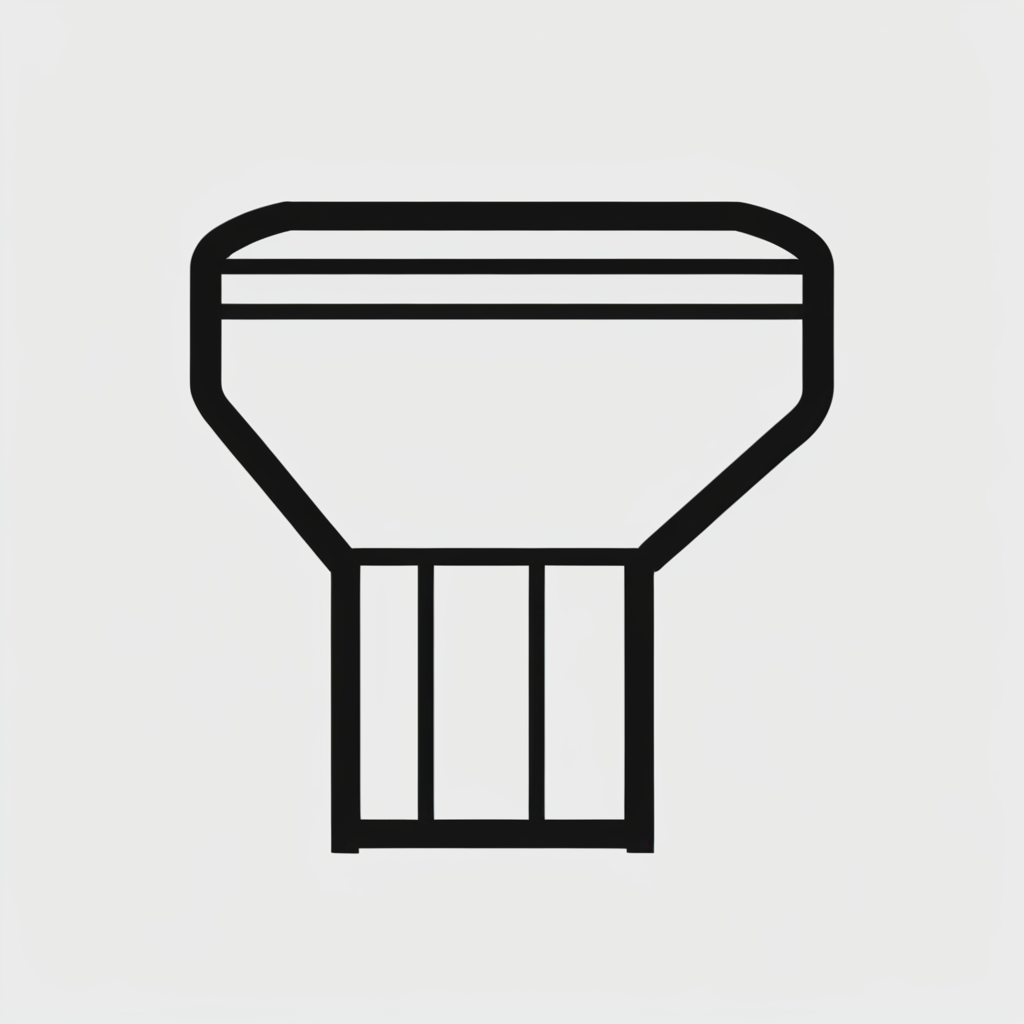}
        \caption{Phillips drive oval head screw icon}
        \label{fig:class_phillips_oval_short}
    \end{subfigure}
    \hfill
    \begin{subfigure}{0.22\textwidth}
        \centering
        \includegraphics[width=\linewidth]{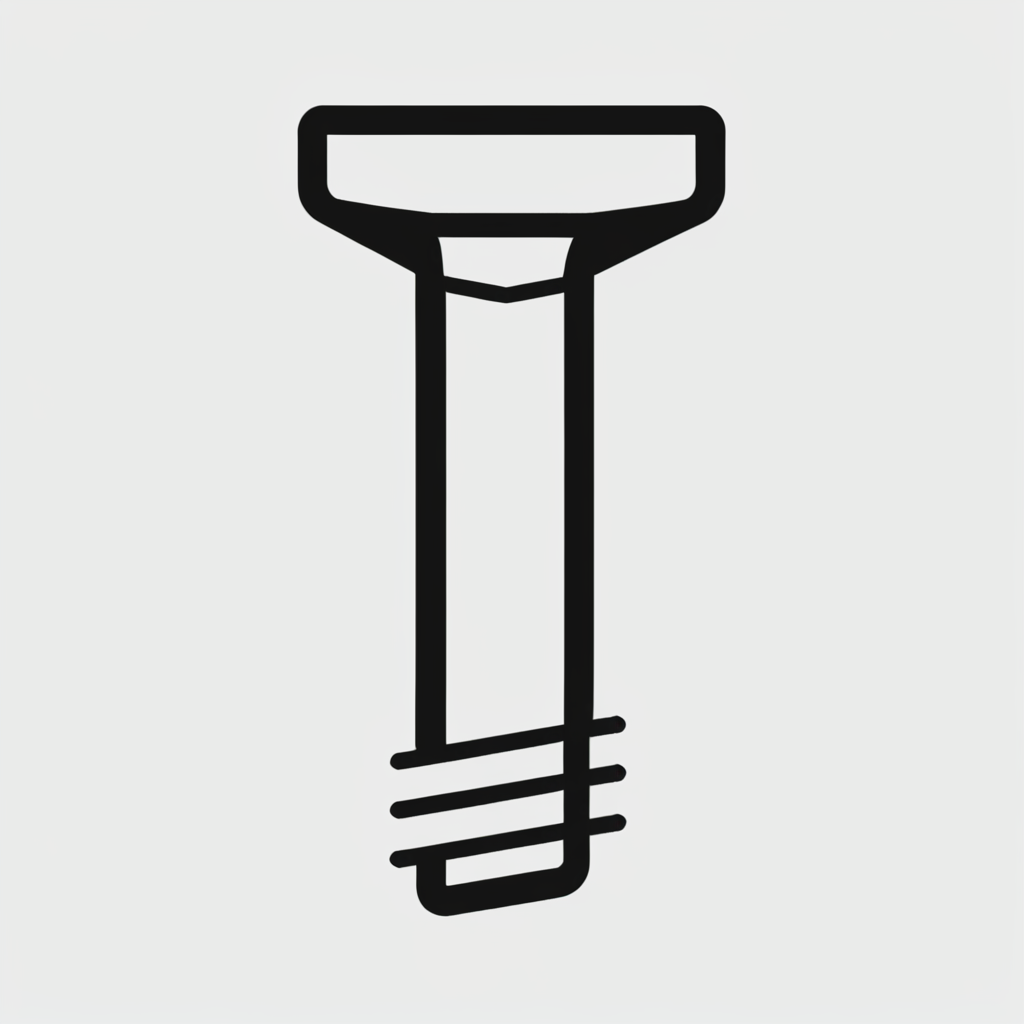}
        \caption{Spanner drive pan head screw icon}
        \label{fig:class_spanner_pan_short}
    \end{subfigure}
    \caption{Example generated icons based on commercial training + short prompt class images (each caption above is what was used at inference time)}
    \label{fig:class_home_depot_short_prompts}
\end{figure}

For the model trained with short prompt class images, this is the model with the best FID score objectively as identified previously in Figure~\ref{screwfidclip}. Although this model does take the lead for the best FID score, in the context of icon generation, we can see that these icons are not the best from the human assessment perspective. Specifically, the oval head screw icon (c) and the spanner drive pan head screw icon (d) do not look coherent at all. This substantiates the fact that although the FID score may be better, it does not necessarily mean the icon generated at inference time is representative of the expected outcome. This is due to the fact that not all of the pixels can be captured in reference to the original image which are of importance when it comes to quality. In this case we can say that the short prompt icon generations in Figure~\ref{fig:home_depot_short_prompts} are more representative of the outcome we are looking for, despite having worse FID. 
\begin{figure}[htbp]
\centerline{\includegraphics[width=0.6\linewidth]{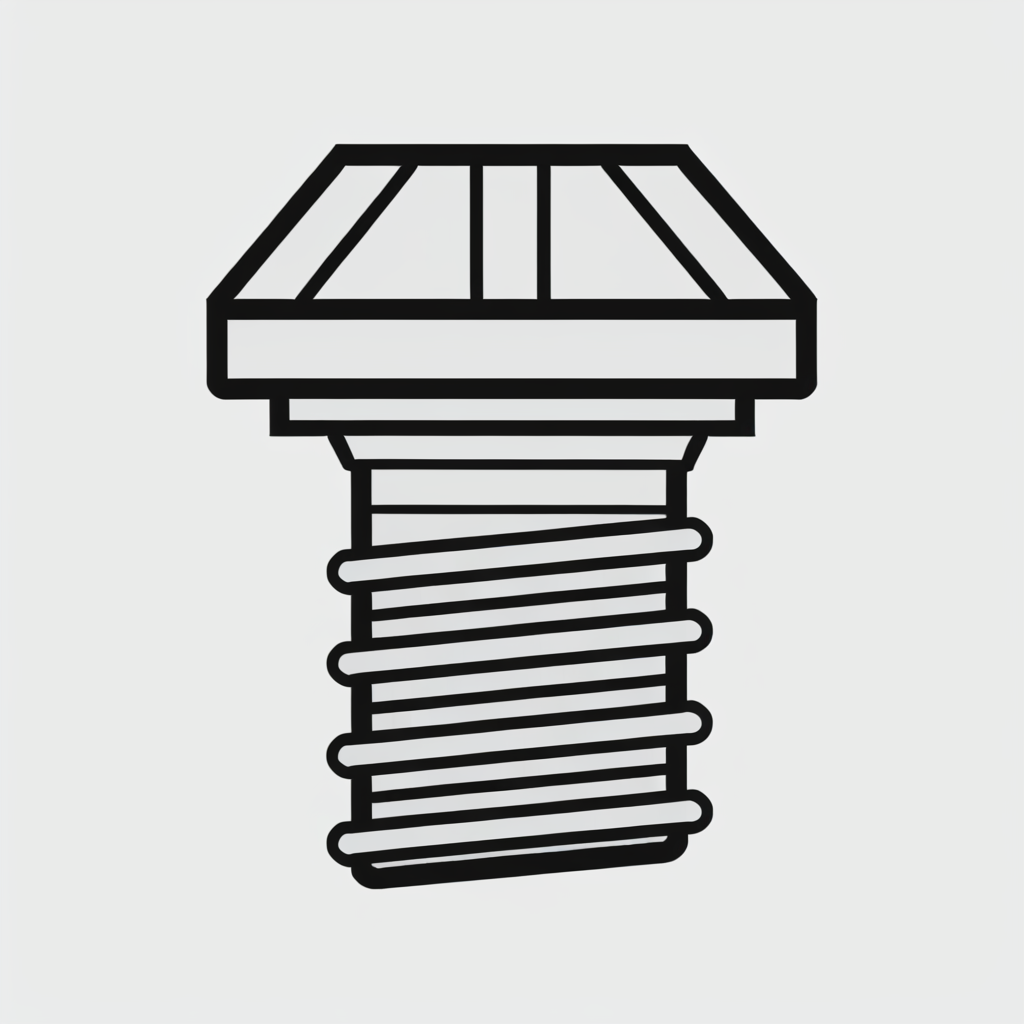}}
\caption{An example of an image with a high clip score (0.30) but incorrect icon generation. (inference: "cross drive round head screw icon")}
\label{longpromptcommercial}
\end{figure}

Another interesting point is that out of the short prompts which curated the highest clip scores out of all (0.30) it is evident that just because clip scores are high, does not mean the expected outcome is correct. In Figure~\ref{longpromptcommercial} we can see that this image has a high clip score in comparison to the other images which are $\leq 0.299$. However, its generation is incorrect. This is because despite the inference prompt to generate the image "cross drive round head screw icon", it generates a screw that is not a cross drive round head, but rather a different head. This substantiates the fact CLIP scores are relative to the training data the CLIP model is based off of, and it is not always able to generalize to nuanced use-cases, such as stylistic icon generation. This further substantiates why human-assessment of use-cases such as these is paramount. 

As a cross comparison, the same prompts are used in DALL-E 3 to understand the FID score and CLIP score as well as look at the results without any training. The scores and generated images can be viewed in the Appendix in Figure~\ref{fig:home_depot_dalle_prompts}. DALL-E 3 shows the highest average CLIP score and the lowest FID score calculated based on the commercial training set for screws. 

\FloatBarrier


\subsection{Models trained on commercial data for Kitchen Cabinets}
\begin{figure}[htbp]
\centerline{\includegraphics[width=1\linewidth]{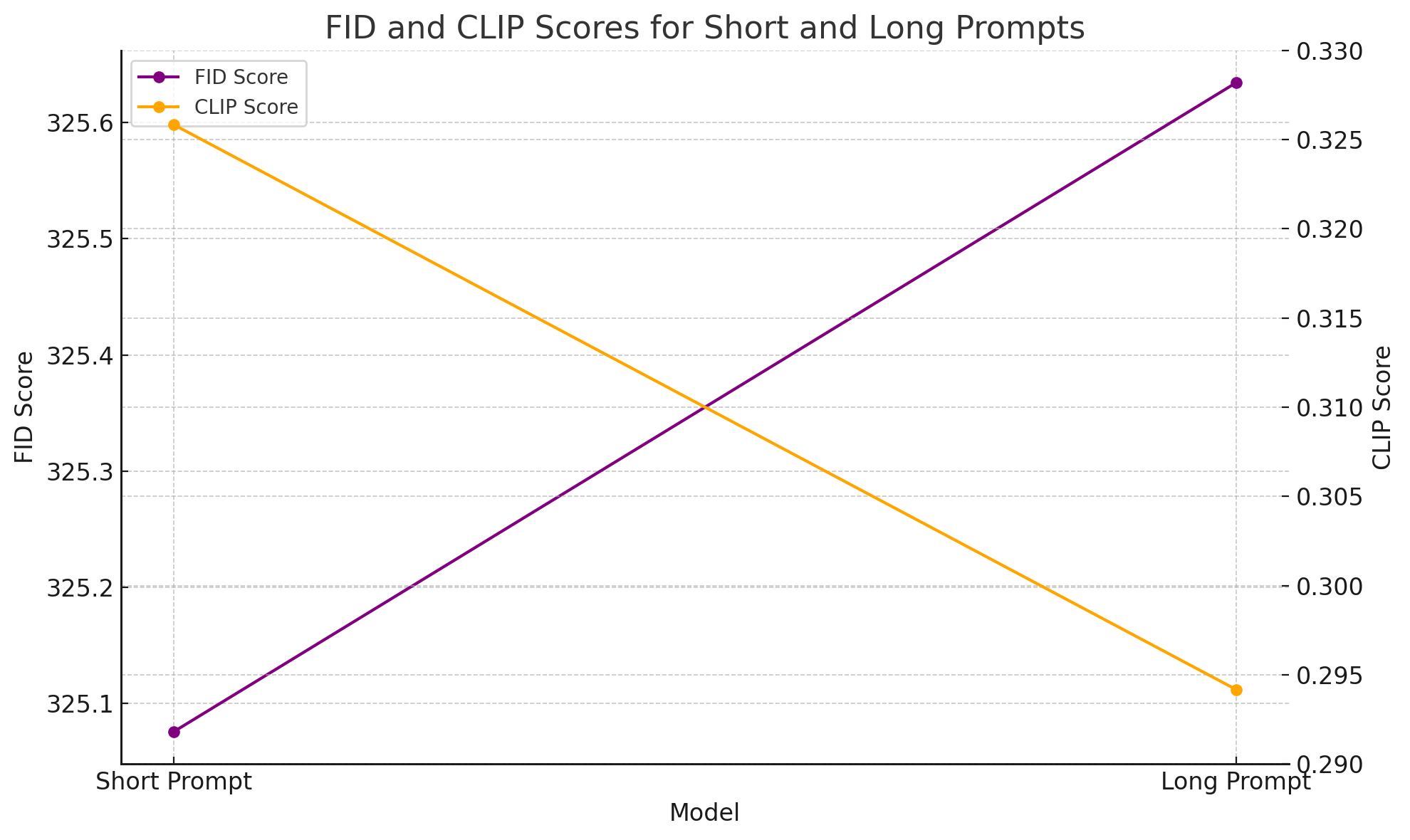}}
\caption{FID and CLIP scores of kitchen cabinet icons generated using commercial icons for training (non-public data)}
\label{kitchenfidclip}
\end{figure}
Using the model trained on commercial data with longer captions and shorter captions for kitchen cabinet icon generation, the FID and CLIP scores can be viewed in Figure~\ref{kitchenfidclip}. Here it can be seen that the short prompt images perform the best in terms of CLIP and FID. The short prompt with class images was not conducted for kitchen cabinets because it was highly computational, and the performance with the long prompt was worse than not using the class. Therefore, it was not included in this experiment.
\FloatBarrier

\begin{figure}[htbp]
    \centering
    \begin{subfigure}{0.2\textwidth}
        \centering
        \includegraphics[width=\linewidth]{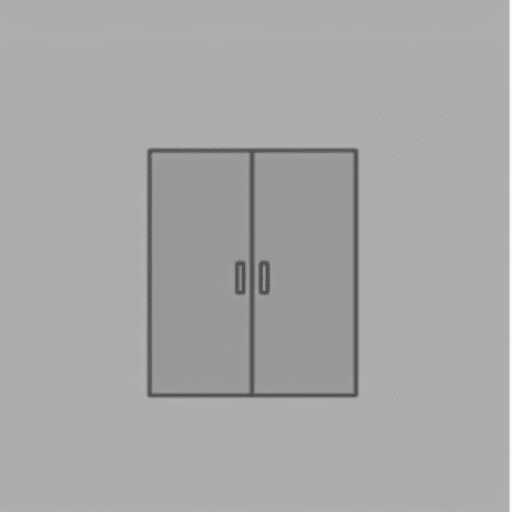}
        \caption{\textbf{Short Prompt}: $<$style: 2D icon$>$, $<$cabinet type: wall$>$, $<$color: gray$>$}
        \label{fig:wall_kitchen_cabinet_grey_short}
    \end{subfigure}
    \hfill
    \begin{subfigure}{0.2\textwidth}
        \centering
        \includegraphics[width=\linewidth]{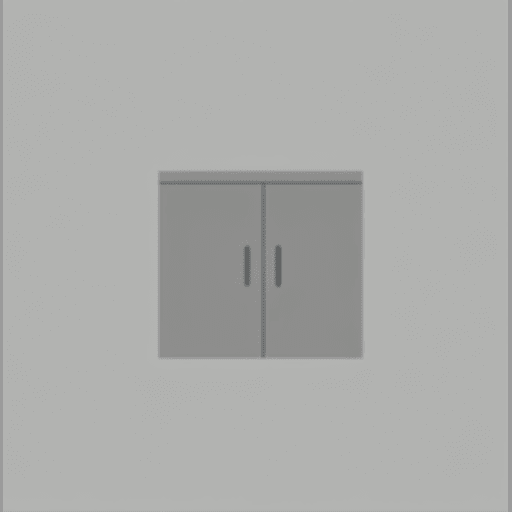}
        \caption{\textbf{Long Prompt}: a photo of TOK kitchen cabinet icon, $<$style: 2D icon$>$, a wall kitchen cabinet in grey with two doors}
        \label{fig:wall_kitchen_cabinet_grey_long}
    \end{subfigure}
    \hfill
    \begin{subfigure}{0.2\textwidth}
        \centering
        \includegraphics[width=\linewidth]{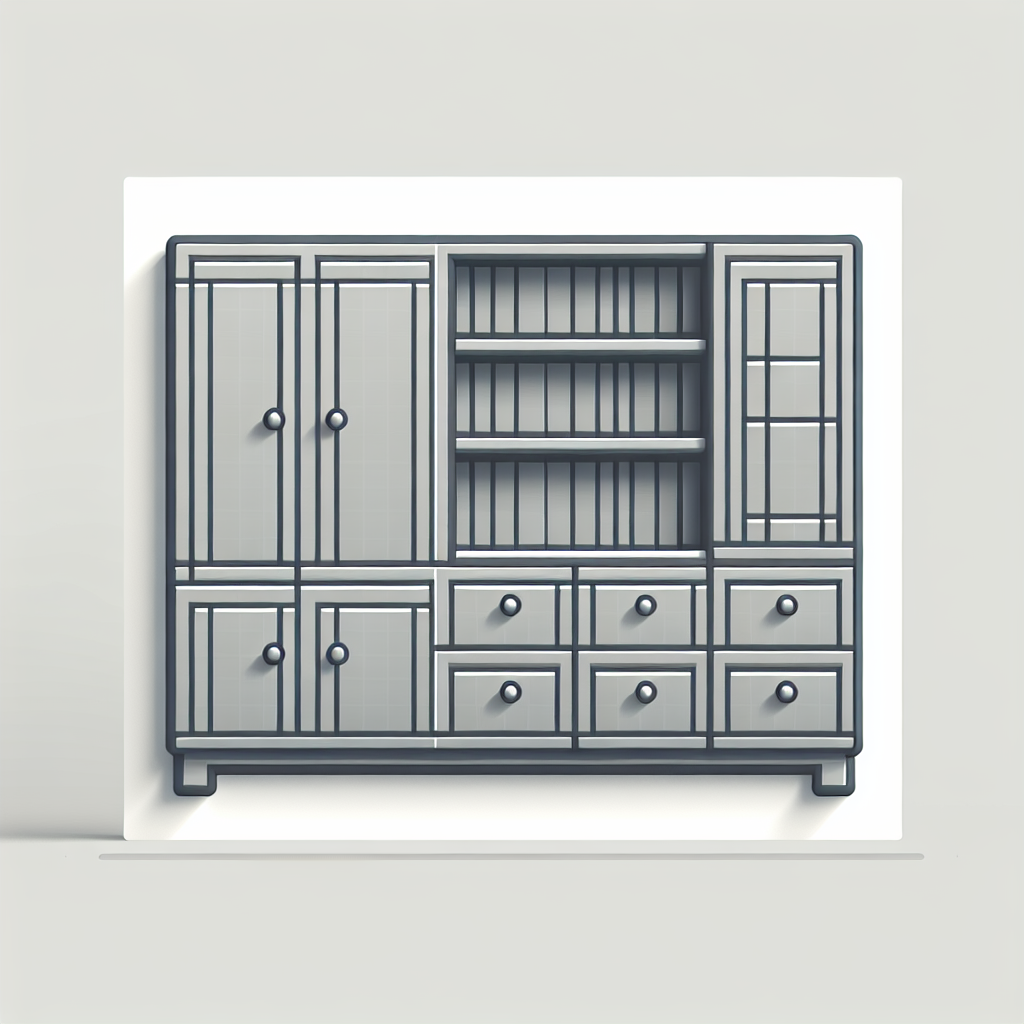}
        \caption{\textbf{DALL-E 3 short prompt}: $<$style: 2D icon$>$, $<$cabinet type: wall$>$, $<$color: gray$>$}
        \label{fig:wall_kitchen_cabinet_grey_dalle}
    \end{subfigure}
    \caption{Wall Kitchen Cabinet in Grey}
    \label{fig:wall_kitchen_cabinet_grey}
\end{figure}
\FloatBarrier

\begin{figure}
\centering
\begin{subfigure}{0.2\textwidth}
    \includegraphics[width=\linewidth]{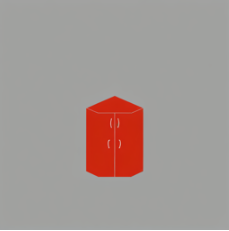}
    \caption{\textbf{Short Prompt}: $\langle style:\ 3D\ icon\rangle$, $\langle cabinet\ type:\ corner\rangle$, $\langle color:\ red\rangle$}
    \label{fig:subim1}
\end{subfigure}%
\hfill
\begin{subfigure}{0.2\textwidth}
    \includegraphics[width=\linewidth]{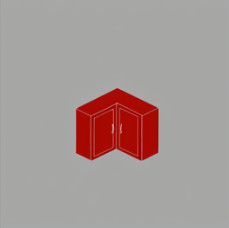}
    \caption{\textbf{Long Prompt}: a photo of TOK kitchen cabinet icon, $\langle style:\ 3D\ icon\rangle$, a corner kitchen cabinet in red with doors at the bottom}
    \label{fig:subicm2}
\end{subfigure}%
\hfill
\begin{subfigure}{0.2\textwidth}
    \includegraphics[width=\linewidth]{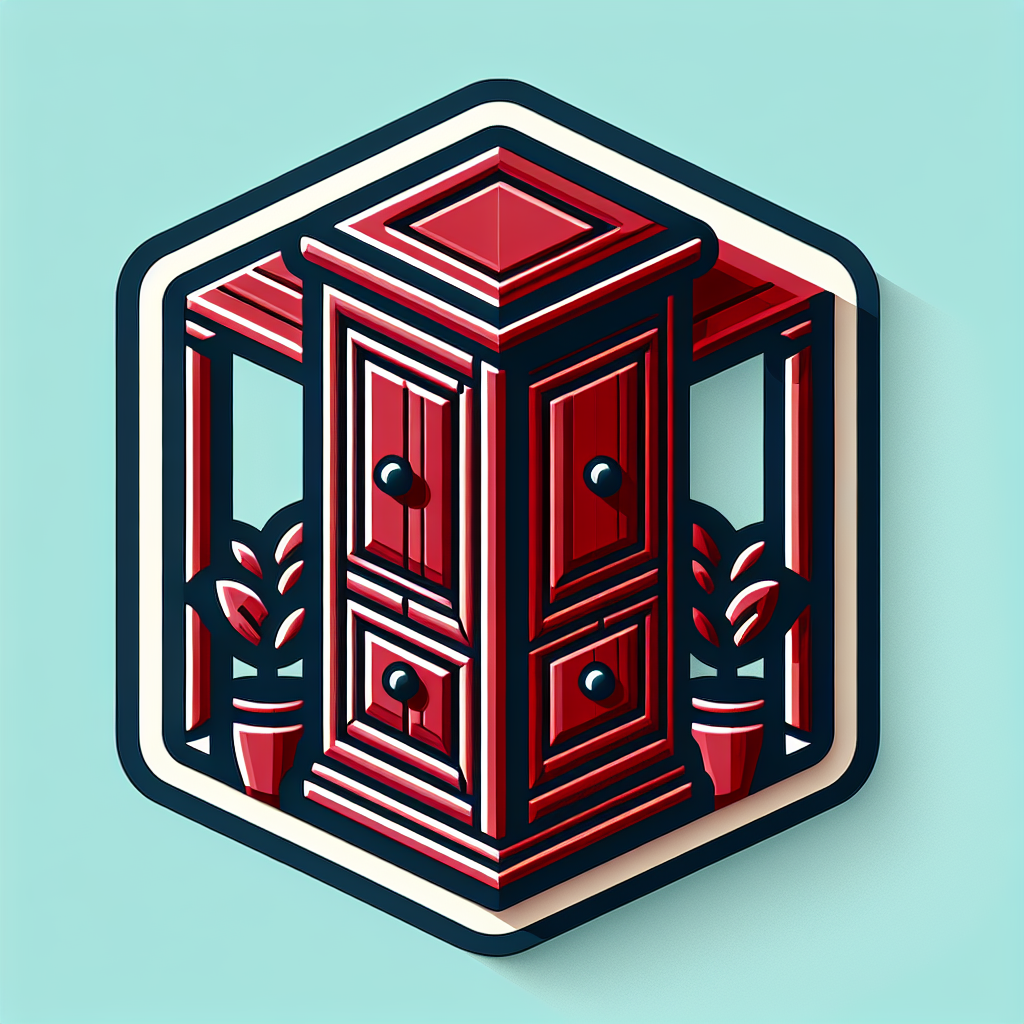}
    \caption{\textbf{DALL-E 3 short prompt}:$\langle style:\ 3D\ icon\rangle$, $\langle cabinet\ type:\ corner\rangle$, $\langle color:\ red\rangle$}
    \label{fig:subim1}
\end{subfigure}
\caption{Corner Kitchen Cabinet in Red}
\end{figure}

\begin{figure}[htbp]
\centerline{\includegraphics[width=.89\linewidth]{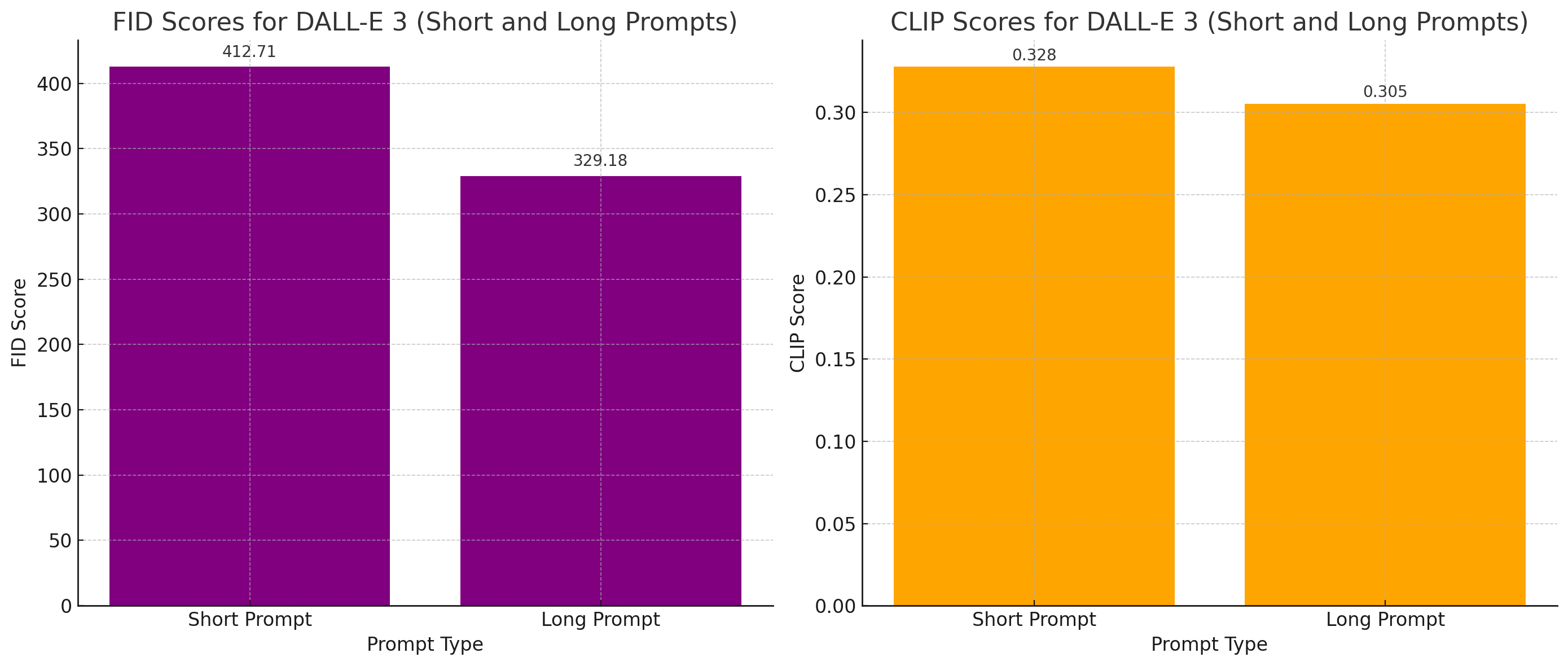}}
\caption{FID and CLIP scores of kitchen icons generated using DALL-E 3}
\label{dallefidclipkitchen}
\end{figure}

\begin{figure}
\centering
\begin{subfigure}{0.2\textwidth}
    \includegraphics[width=\linewidth]{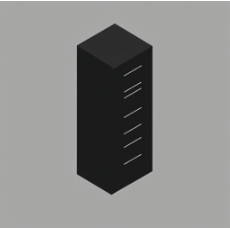}
    \caption{\textbf{Short Prompt}: $\langle style:\ 3D\ icon\rangle$, $\langle cabinet\ type:\ drawer\ base\rangle$, $\langle color:\ black\rangle$}
    \label{fig:subim1}
\end{subfigure}%
\hfill
\begin{subfigure}{0.2\textwidth}
    \includegraphics[width=\linewidth]{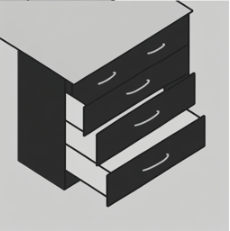}
    \caption{\textbf{Long Prompt}: a photo of TOK kitchen cabinet icon, $\langle style:\ 3D\ icon\rangle$, a drawer base kitchen cabinet in black with three drawers}
    \label{fig:subim2f}
\end{subfigure}%
\hfill
\begin{subfigure}{0.2\textwidth}
    \includegraphics[width=\linewidth]{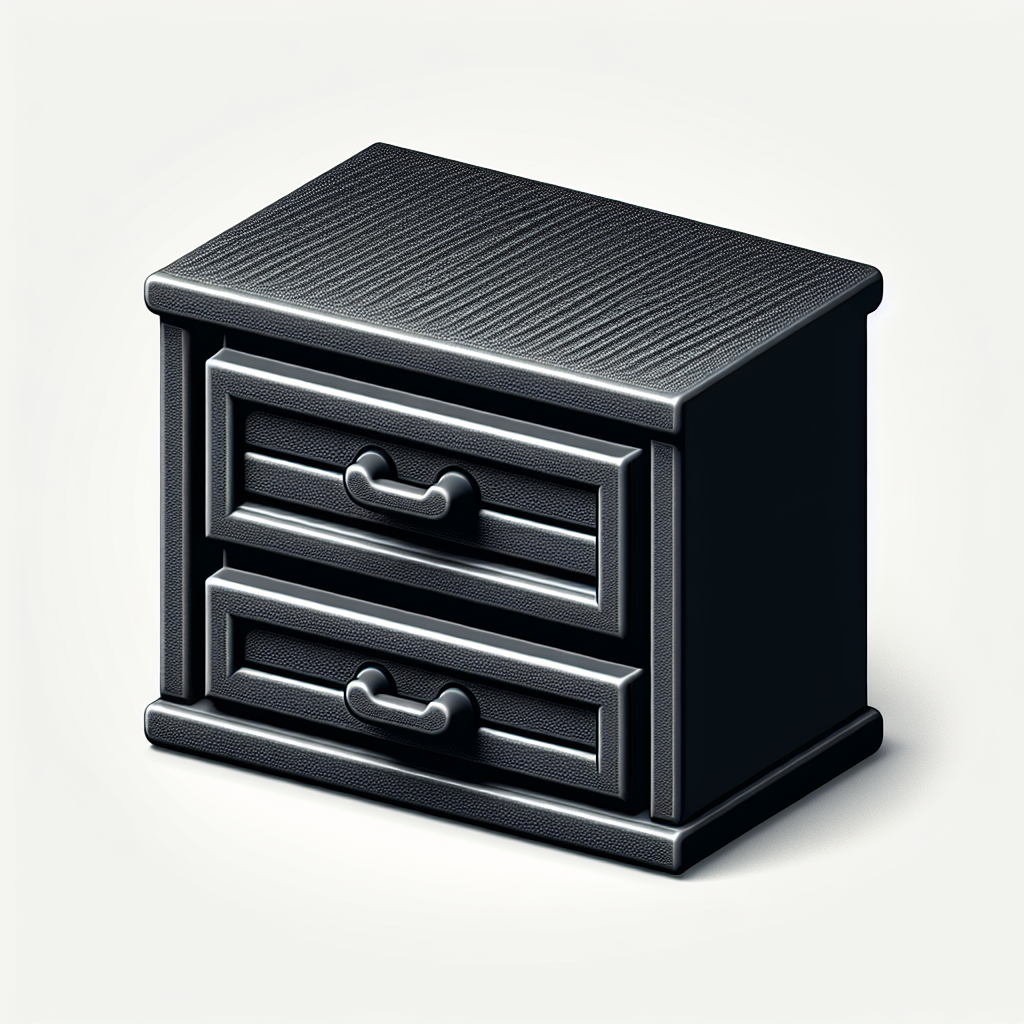}
    \caption{\textbf{DALL-E 3 short prompt}:$\langle style:\ 3D\ icon\rangle$, $\langle cabinet\ type:\ drawer\ base\rangle$, $\langle color:\ black\rangle$}
    \label{fig:subim3}
\end{subfigure}
    
\caption{Drawer Base Kitchen Cabinet in Black}
\end{figure}

Similar to the findings from screw icons, the short prompt trained model has a higher CLIP score. However, when looking at the FID score, the short and long prompt have very similar FID scores as seen in Figure~\ref{kitchenfidclip}. DALL-E 3 has a better CLIP score (see Figure~\ref{dallefidclipkitchen}), but this does not necessarily mean that the icon is what we are looking for, which makes sense. Although the short prompt trained model has better performance than the long prompt model, when looking at CLIP and FID scores, there are cases where long prompt generated icons are visually better and more customized based on the description in the prompt. This can be seen in Figure~\ref{fig:subicm2} and Figure~\ref{fig:subim2f}. This further substantiates the fact that a high CLIP score is not a direct measure of whether a generated icon is in the correct style.

\begin{figure}
\centering
\begin{subfigure}{0.2\textwidth}
    \includegraphics[width=\linewidth]{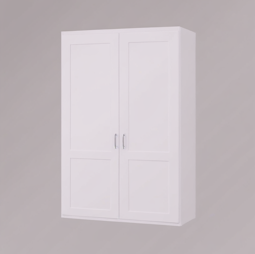}
    \caption{\textbf{Long Prompt}: A photo of TOK kitchen cabinet icon, photo of a standard wall kitchen cabinet in white with two doors}
    \label{fig:subim5}
\end{subfigure}%
\hfill
\begin{subfigure}{0.2\textwidth}
    \includegraphics[width=\linewidth]{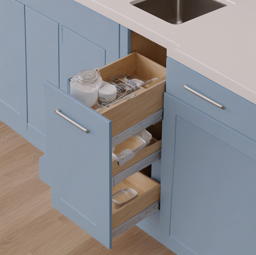}
    \caption{\textbf{Long Prompt}: A photo of TOK kitchen cabinet icon, photo of a standard kitchen cabinet in blue with pull out drawers and sink on top}
    \label{fig:subim6}
\end{subfigure}%
\hfill
\begin{subfigure}{0.2\textwidth}
    \includegraphics[width=\linewidth]{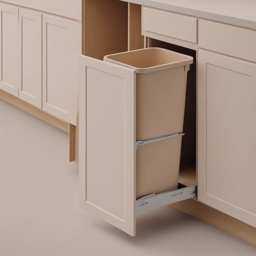}
    \caption{\textbf{Long prompt}: A photo of TOK kitchen cabinet icon, photo of a trash can kitchen cabinet in unfinished color with a pull-out trash bin}
    \label{fig:subim7}
\end{subfigure}
\caption{Kitchen Cabinet with realistic product photos as Training Data}
\end{figure}

Additionally, when comparing results from Stable Diffusion XL with DALL-E 3, we observe a noticeable difference in style. This highlights the importance of training data in generating customized and consistent styles. To explore this further, we trained several models with a combination of icons and realistic product photos as the training data and included style variable in the caption jsonl file. The results show that incorporating even a few realistic product photos can significantly skew the generated icon style to match that of the realistic photos. This can be seen in Figures~\ref{fig:subim7},~\ref{fig:subim6},~\ref{fig:subim5}.

\FloatBarrier


\FloatBarrier

\section{Discussion}
\subsection{Key Findings}
In this study, we fine-tuned Stable Diffusion XL with different training data and caption sizes to generate stylistic icons for a commercial environment. Our focus was on how different training prompts and evaluation metrics impact the quality and relevance of generated icons. We found that short prompts with class images yielded the best FID scores but often failed to meet qualitative expectations when assessed by humans. Long prompts, though slightly higher in FID scores, produced icons that better aligned with the intended style surprisingly in some cases. This highlights the importance of human assessment in evaluating the true quality of generated icons.
While FID scores measure overall similarity between generated images and the training set, they miss significant aspects such as detailed stylistic features. High CLIP scores were found to not always correlate with the expected visual outcome based on a generated image of the wrong screw that had the highest CLIP score out of all. Also, we observed that DALL-E 3 generally had higher average CLIP scores and FID scores. However, DALL-E 3’s generated icons were often less customized to the specific styles we aimed for. Fine-tuning Stable Diffusion XL with detailed prompts showed better consistency and accuracy in style sometimes for short and sometimes for long captions with varying degree. Our experiments also revealed that incorporating realistic product photos in the training set skewed the generated icons towards a more realistic style, which was not desirable for our use case of 2D stylistic icons. 
\subsection{Limitations}
Our study utilized relatively small datasets - 42 screw icons and 42 kitchen cabinet icons. This limited sample size may not fully represent the diversity of icons in these categories, potentially affecting the generalizability of our findings.
We focused on only two types of icons (screws and kitchen cabinets). While these provided interesting contrasts, they represent a narrow slice of the vast array of icon types used in commercial settings. Our findings may not apply equally to all icon categories.
Due to limited computational resources, we were unable to fully explore certain aspects of the model, such as finding the optimal prior\_loss\_weight for class images. Additionally, while we used FID and CLIP scores as evaluation metrics, these have limitations in capturing the nuanced stylistic elements crucial for icon quality.
\subsection{Implications and Future Directions}
Our findings have significant implications for the commercial application of AI-generated imagery, particularly in contexts where style consistency is crucial. The discrepancy between quantitative metrics and human assessment underscores the need for more sophisticated evaluation methods in AI-generated art.

We conclude that CLIP score and FID score are insufficient alone for icon generation. Tailored evaluation criteria based on specific business use-cases, supplemented by human assessments, are crucial for ensuring the generated icons meet the desired quality and style. Future work should focus on developing refined evaluation metrics that better capture stylistic nuances and quality requirements from an automation perspective. For example, exploring larger and more diverse datasets to improve the generalizability of the fine-tuning process, studying the balance between stylized and realistic training data and determining optimal training set compositions for different icon styles and use cases.

\section{Appendix}
\subsection{Additional DALL-E 3 Screw Prompt Comparisons}
\begin{figure}[htbp]
    \centering
    \begin{subfigure}{0.22\textwidth}
        \centering
        \includegraphics[width=\linewidth]{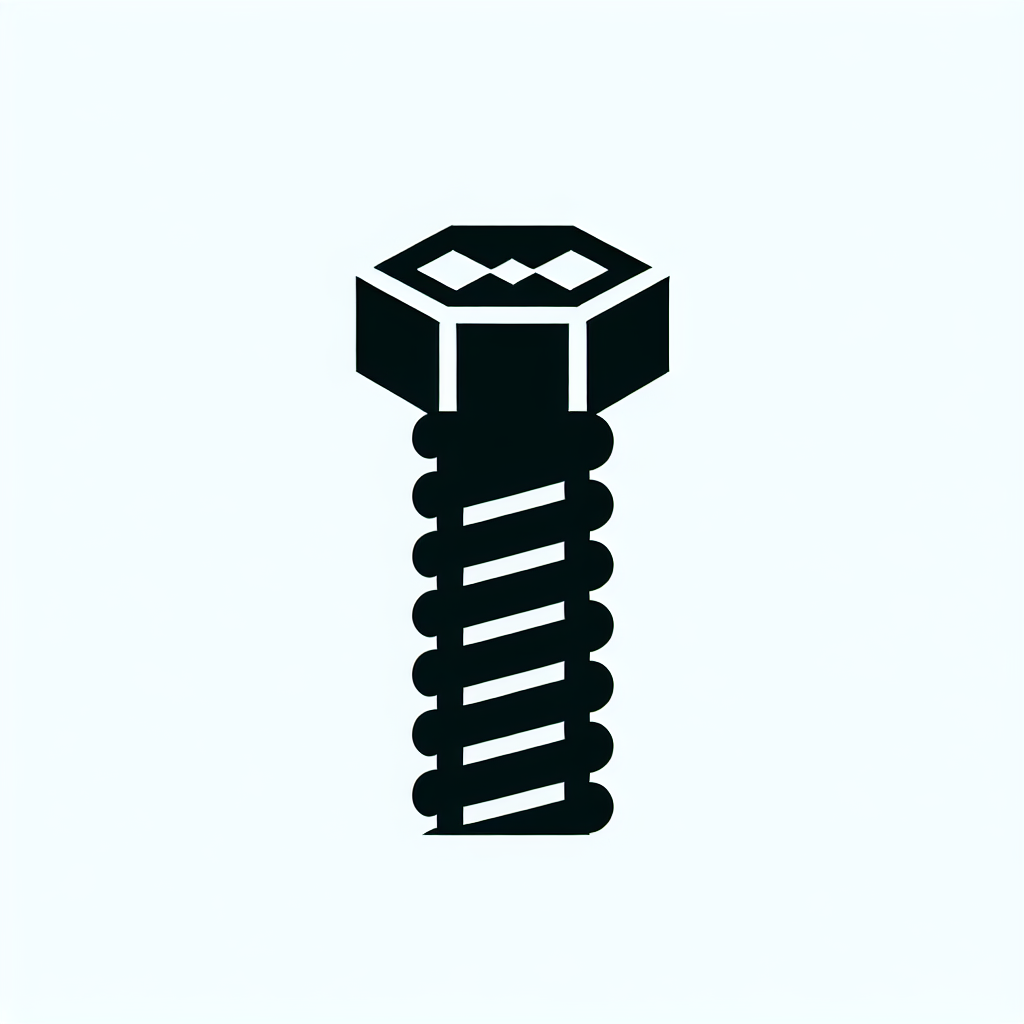}
        \caption{Hex head screw icon}
        \label{fig:hex_head_dalle}
    \end{subfigure}
    \hfill
    \begin{subfigure}{0.22\textwidth}
        \centering
        \includegraphics[width=\linewidth]{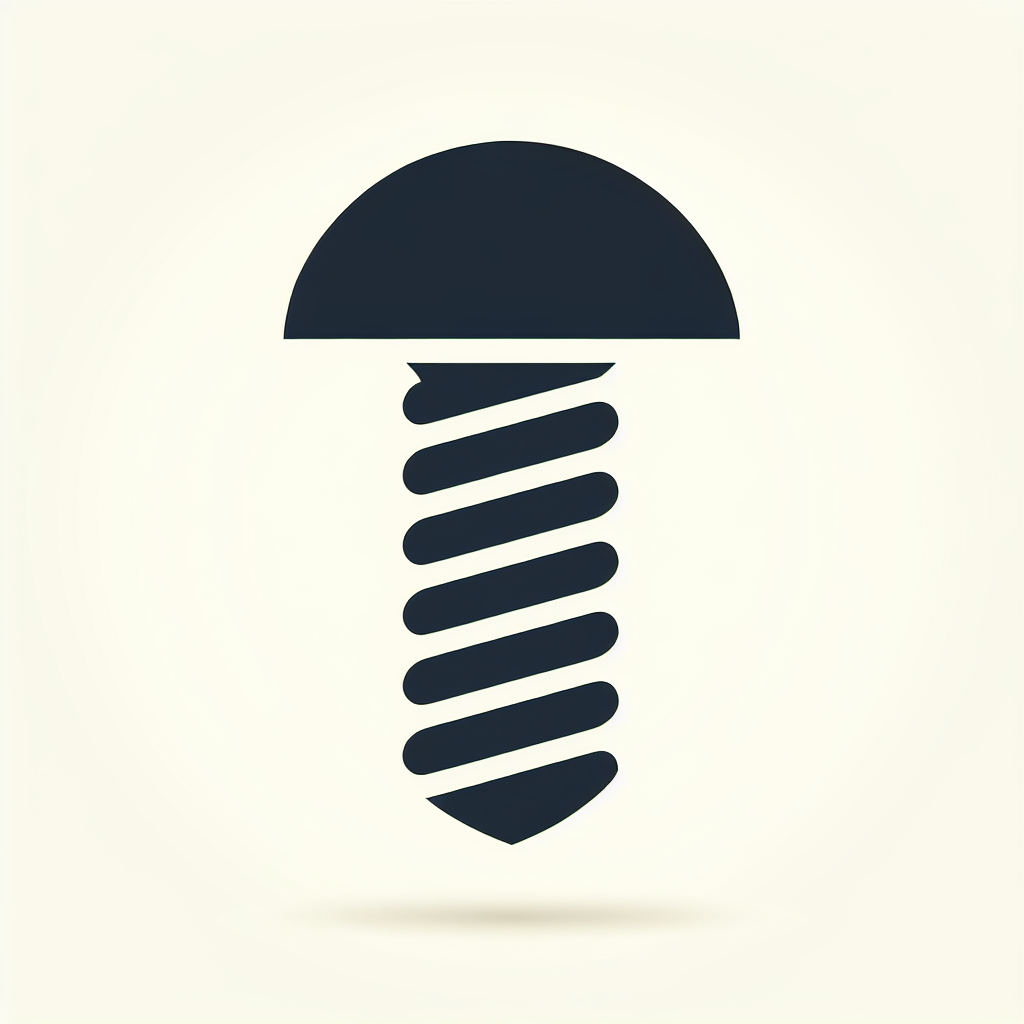}
        \caption{Round head screw icon}
        \label{fig:round_head_dalle}
    \end{subfigure}
    \hfill
    \begin{subfigure}{0.22\textwidth}
        \centering
        \includegraphics[width=\linewidth]{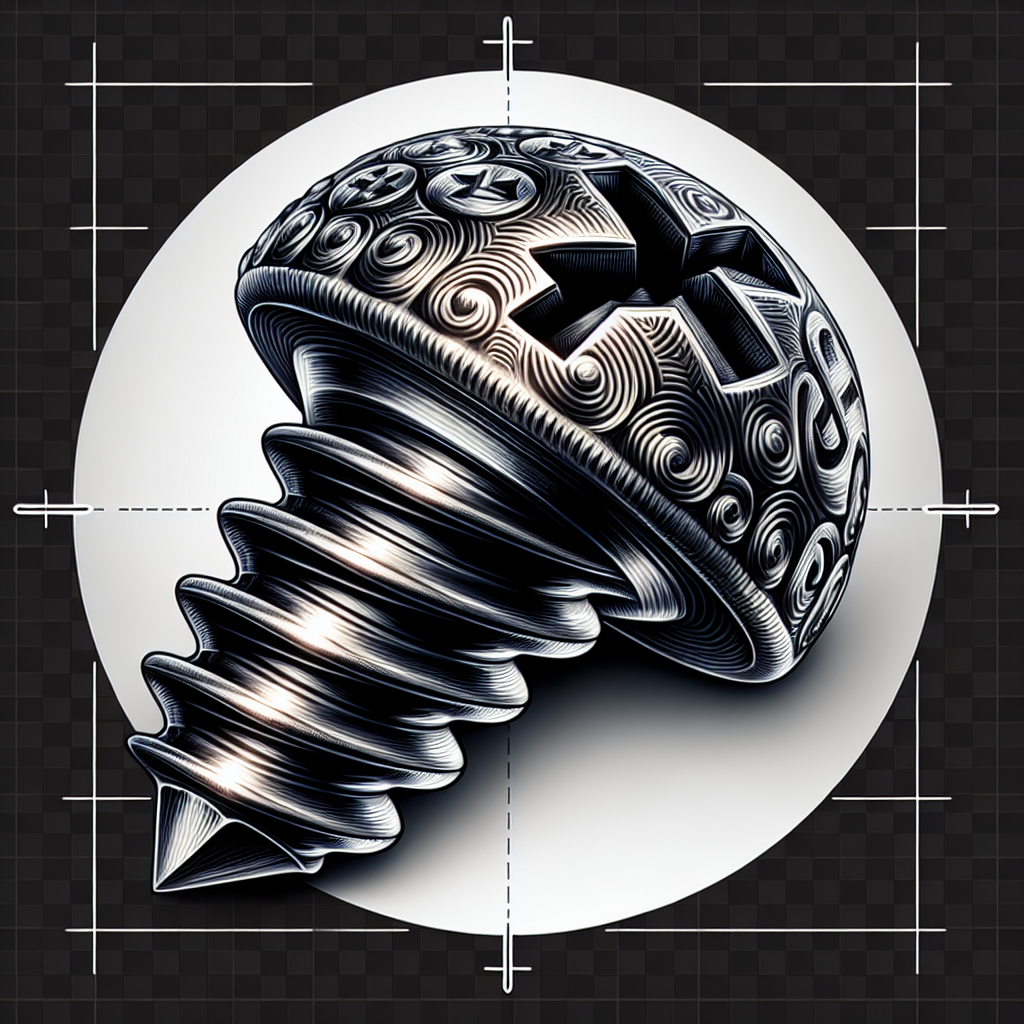}
        \caption{Phillips drive oval head screw icon}
        \label{fig:phillips_oval_dalle}
    \end{subfigure}
    \hfill
    \begin{subfigure}{0.22\textwidth}
        \centering
        \includegraphics[width=\linewidth]{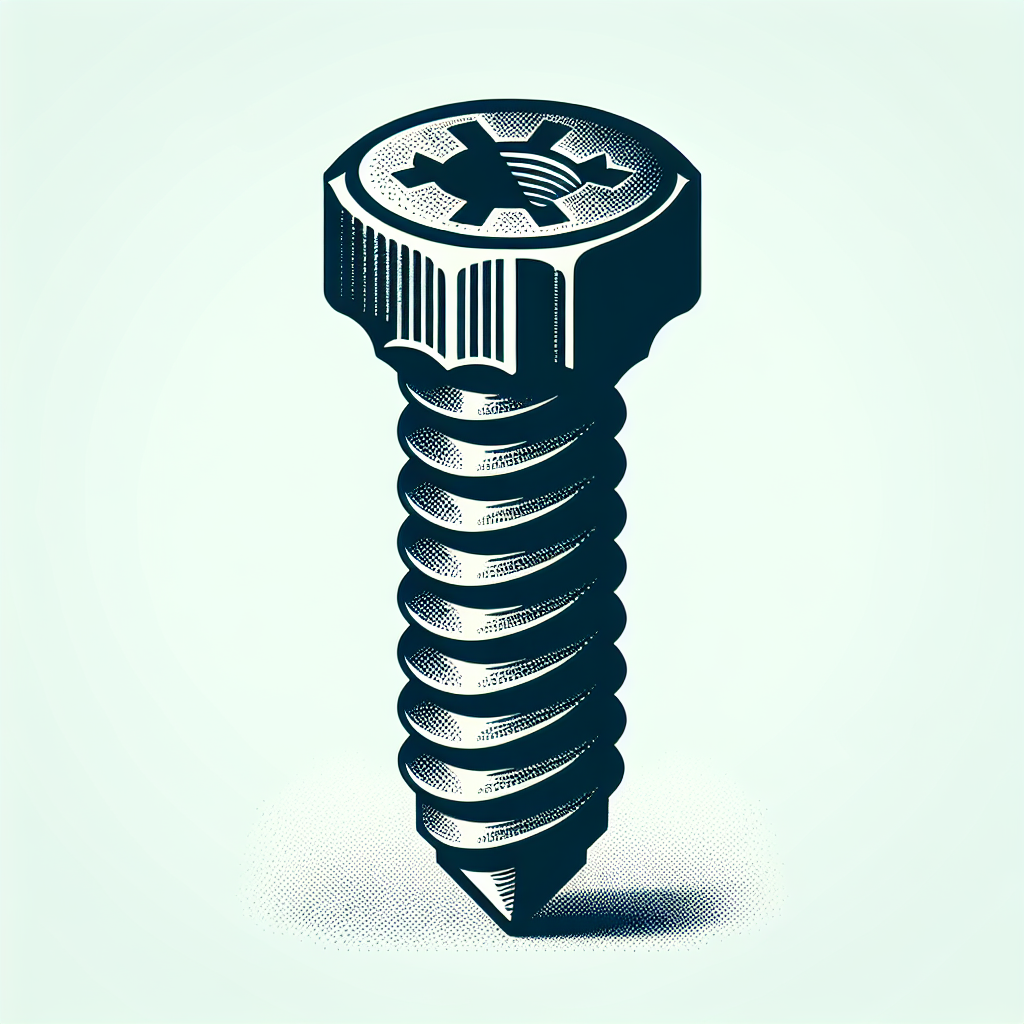}
        \caption{Spanner drive pan head screw icon}
        \label{fig:spanner_pan_dalle}
    \end{subfigure}
    \caption{Screw: Inference prompt generations using DALL-E 3, FID Score: 340, CLIP Score: 0.32}
    \label{fig:home_depot_dalle_prompts}
\end{figure}
\FloatBarrier

\subsection{Additional Kitchen Cabinet Prompt Comparisons}
\begin{figure}[htbp]

    \centering
    \begin{subfigure}{0.2\textwidth}
        \centering
        \includegraphics[width=\linewidth]{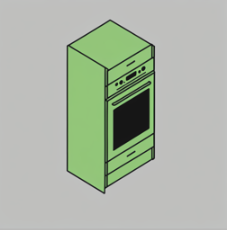}
        \caption{\textbf{Short Prompt}: $<$style: 2D icon$>$, $<$cabinet type: single oven$>$, $<$color: green$>$}
        \label{fig:oven_kitchen_cabinet_green_short}
    \end{subfigure}
    \hfill
    \begin{subfigure}{0.2\textwidth}
        \centering
        \includegraphics[width=\linewidth]{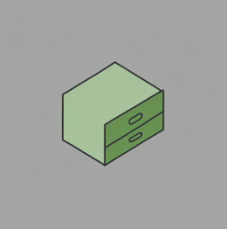}
        \caption{\textbf{Long Prompt}: a photo of TOK kitchen cabinet icon, $<$style: 2D icon$>$, an oven kitchen cabinet in green with one drawer}
        \label{fig:oven_kitchen_cabinet_green_long}
    \end{subfigure}
    \hfill
    \begin{subfigure}{0.2\textwidth}
        \centering
        \includegraphics[width=\linewidth]{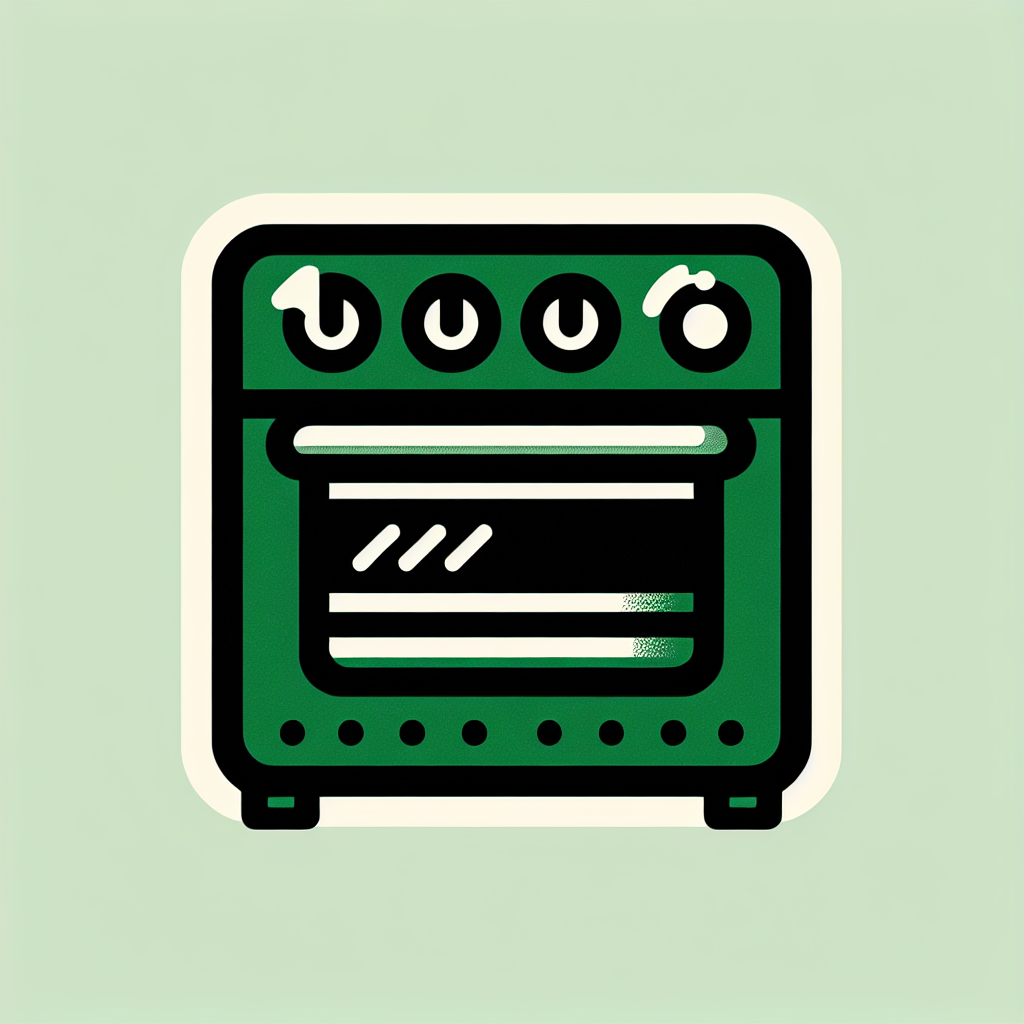}
        \caption{\textbf{DALL-E 3 short prompt}: $<$style: 2D icon$>$, $<$cabinet type: single oven$>$, $<$color: green$>$}
        \label{fig:oven_kitchen_cabinet_green_dalle}
    \end{subfigure}
    \caption{Single Oven Kitchen Cabinet in Green}
    \label{fig:oven_kitchen_cabinet_green}
\end{figure}

\begin{figure}
\centering
\begin{subfigure}{0.2\textwidth}
    \includegraphics[width=\linewidth]{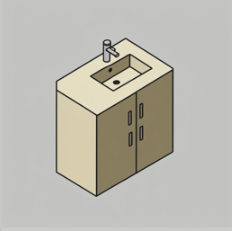}
    \caption{\textbf{Short Prompt}: $\langle style:\ 3D\ icon\rangle$, $\langle cabinet\ type:\ sink\ base\rangle$, $\langle color:\ beige\rangle$}
    \label{fig:subim1}
\end{subfigure}%
\hfill
\begin{subfigure}{0.2\textwidth}
    \includegraphics[width=\linewidth]{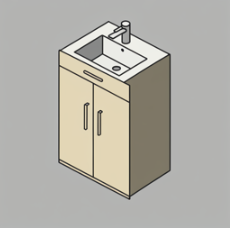}
    \caption{\textbf{Long Prompt}: a photo of TOK kitchen cabinet icon, $\langle style:\ 3D\ icon\rangle$, a sink base kitchen cabinet in beige with a sink on top and two doors at the bottom}
    \label{fig:subim2}
\end{subfigure}%
\hfill
\begin{subfigure}{0.2\textwidth}
    \includegraphics[width=\linewidth]{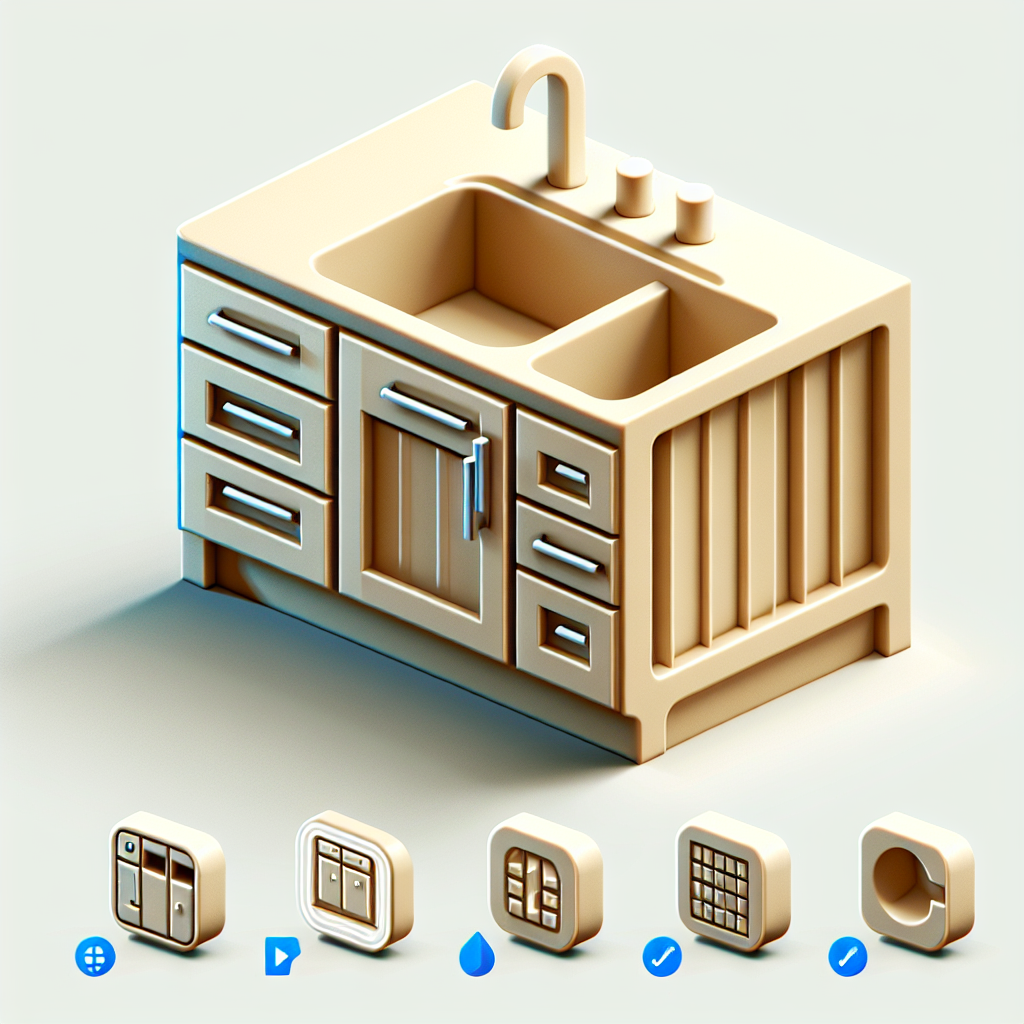}
    \caption{\textbf{DALL-E 3 short prompt}:$\langle style:\ 3D\ icon\rangle$, $\langle cabinet\ type:\ sink\ base\rangle$, $\langle color:\ beige\rangle$}
    \label{fig:subim3}
\end{subfigure}
    
\caption{Sink Base Kitchen Cabinet in Beige}
\end{figure}

\begin{figure}
\centering
\begin{subfigure}{0.2\textwidth}
    \includegraphics[width=\linewidth]{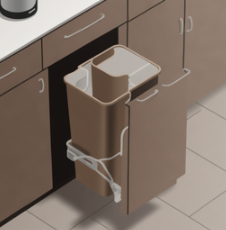}
    \caption{\textbf{Short Prompt}: $\langle style:\ 3D\ icon\rangle$, $\langle cabinet\ type:\ base\rangle$, $\langle color:\ brown\rangle$, $\langle description:\ pull-out\ trash\ can\rangle$}
    \label{fig:subim2}
\end{subfigure}%
\hfill
\begin{subfigure}{0.2\textwidth}
    \includegraphics[width=\linewidth]{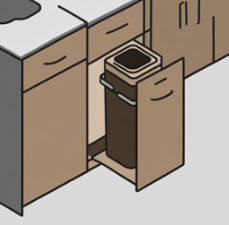}
    \caption{\textbf{Long Prompt}: a photo of TOK kitchen cabinet icon, $\langle style:\ 3D\ icon\rangle$, a kitchen cabinet in brown with pull-out trash can}
    \label{fig:subim3}
\end{subfigure}%
\hfill
\begin{subfigure}{0.2\textwidth}
    \includegraphics[width=\linewidth]{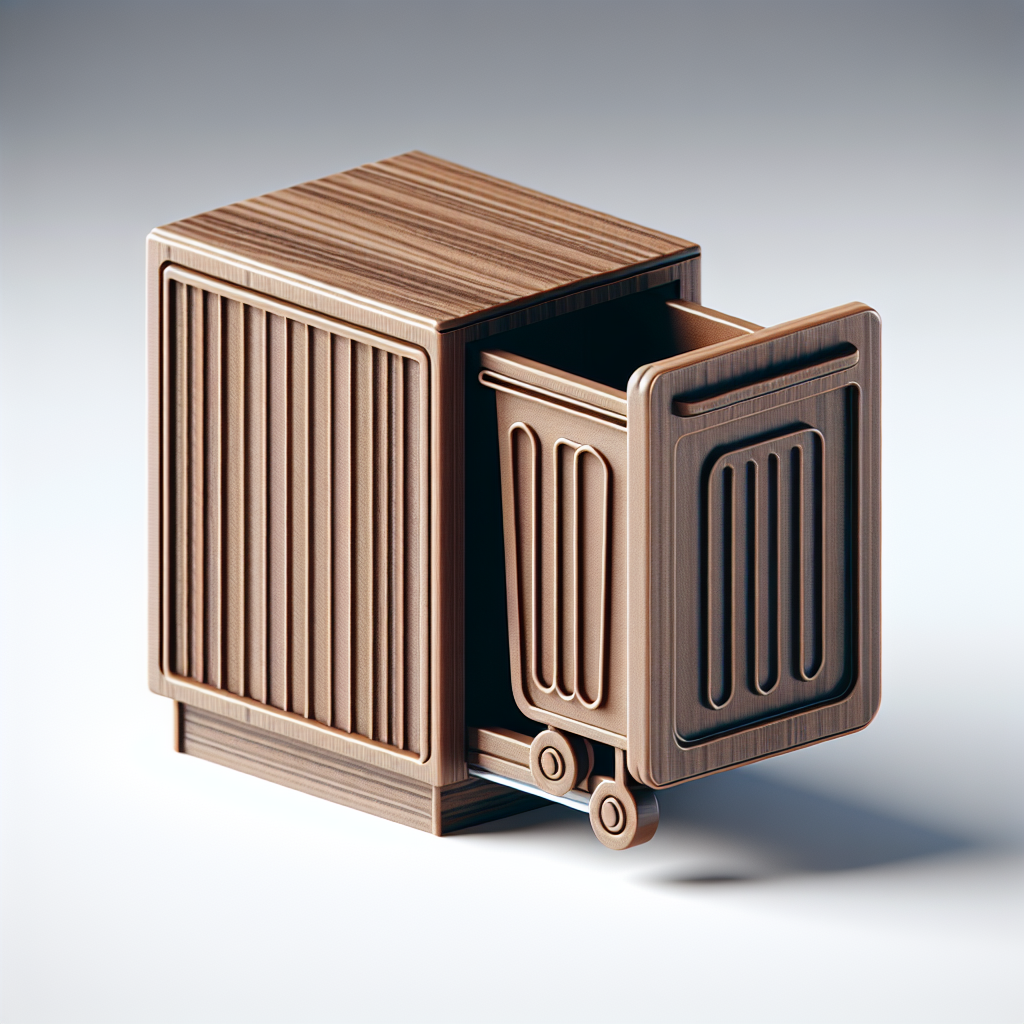}
    \caption{\textbf{DALL-E 3 short prompt}:$\langle style:\ 3D\ icon\rangle$, $\langle cabinet\ type:\ base\rangle$, $\langle color:\ brown\rangle$, $\langle description:\ pull-out\ trash\ can\rangle$}
    \label{fig:subim4}
\end{subfigure}
\caption{Trash Can Kitchen Cabinet in Brown}
\end{figure}

\FloatBarrier

\subsection{Models trained on non-commercial data for Screws}
\begin{figure}[H]
    \centering
    \begin{subfigure}{0.22\textwidth}
        \centering
        \includegraphics[width=\linewidth]{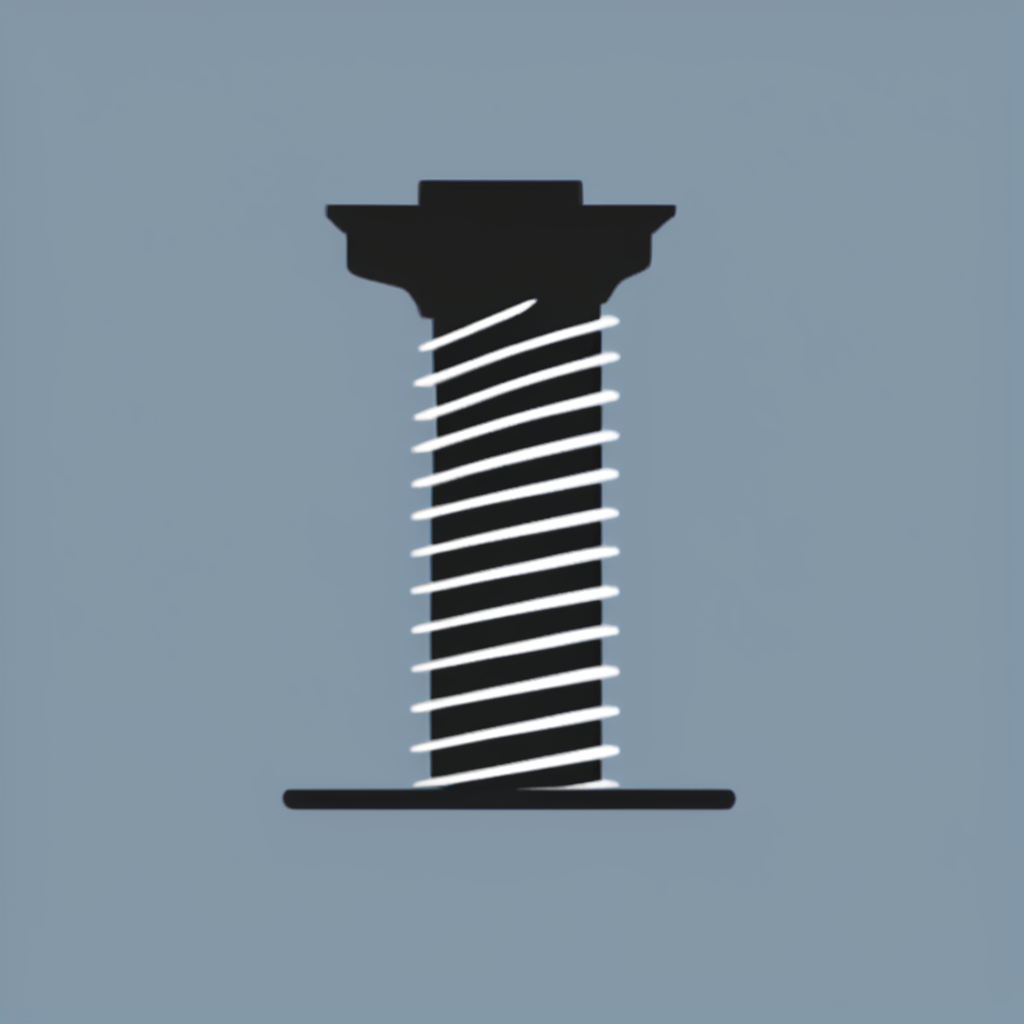}
        \caption{Hex head screw icon}
        \label{fig:public_long_hex_head}
    \end{subfigure}
    \hfill
    \begin{subfigure}{0.22\textwidth}
        \centering
        \includegraphics[width=\linewidth]{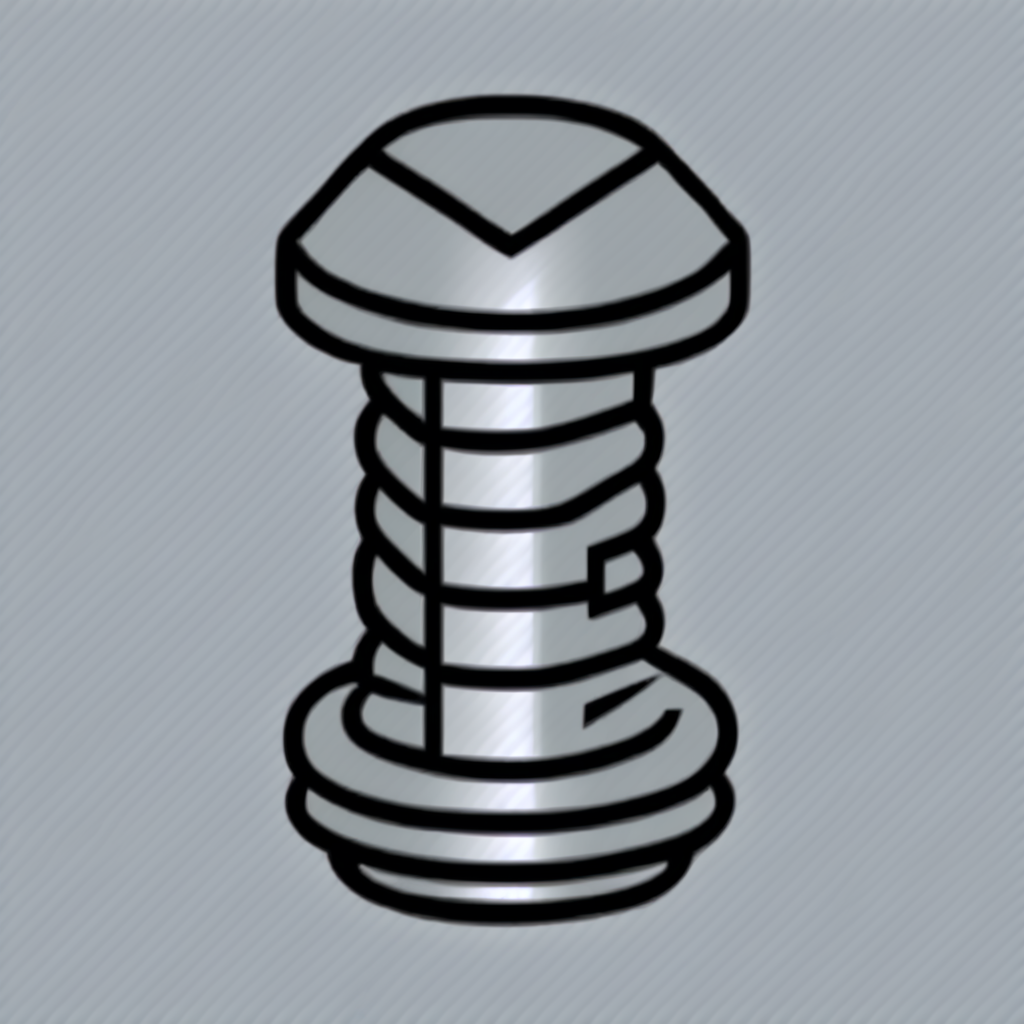}
        \caption{Round head screw icon}
        \label{fig:public_long_round_head}
    \end{subfigure}
    \hfill
    \begin{subfigure}{0.22\textwidth}
        \centering
        \includegraphics[width=\linewidth]{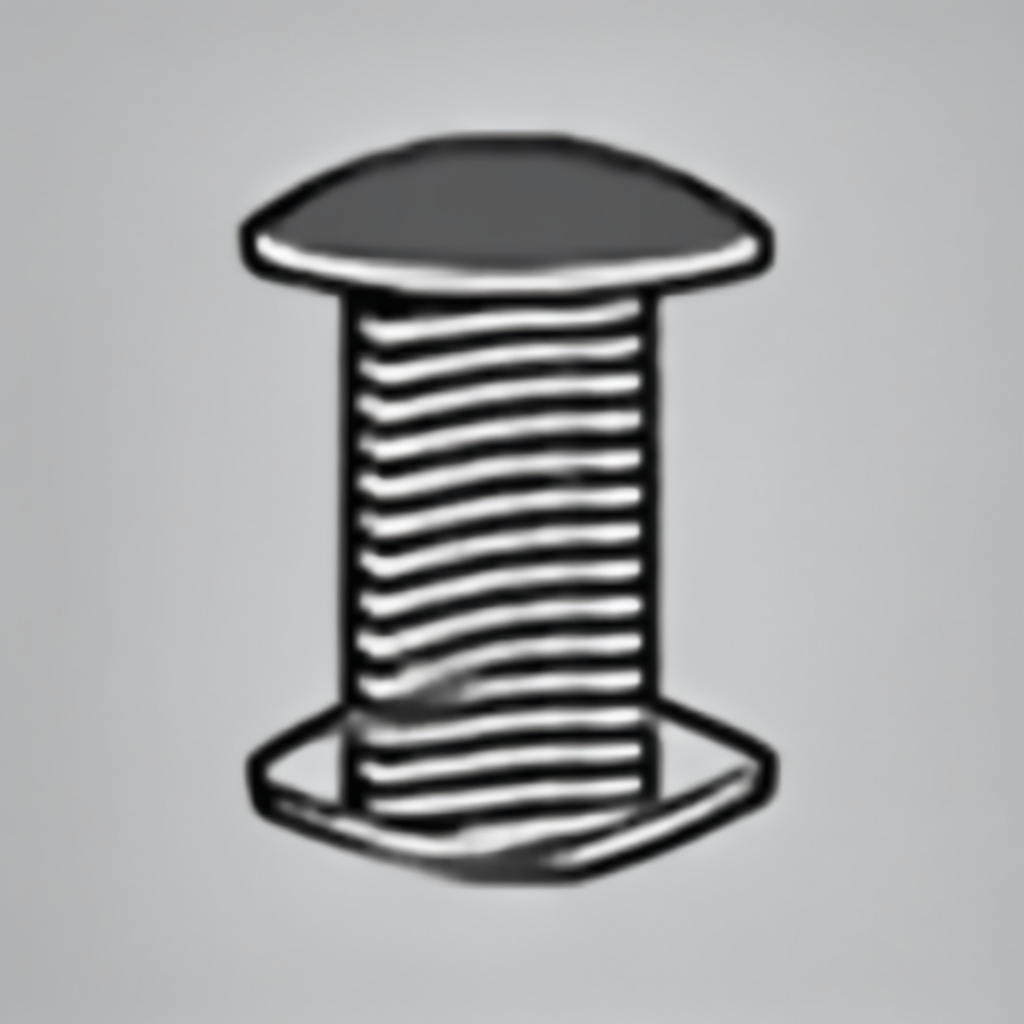}
        \caption{Phillips drive oval head screw icon}
        \label{fig:public_long_phillips_oval}
    \end{subfigure}
    \hfill
    \begin{subfigure}{0.22\textwidth}
        \centering
        \includegraphics[width=\linewidth]{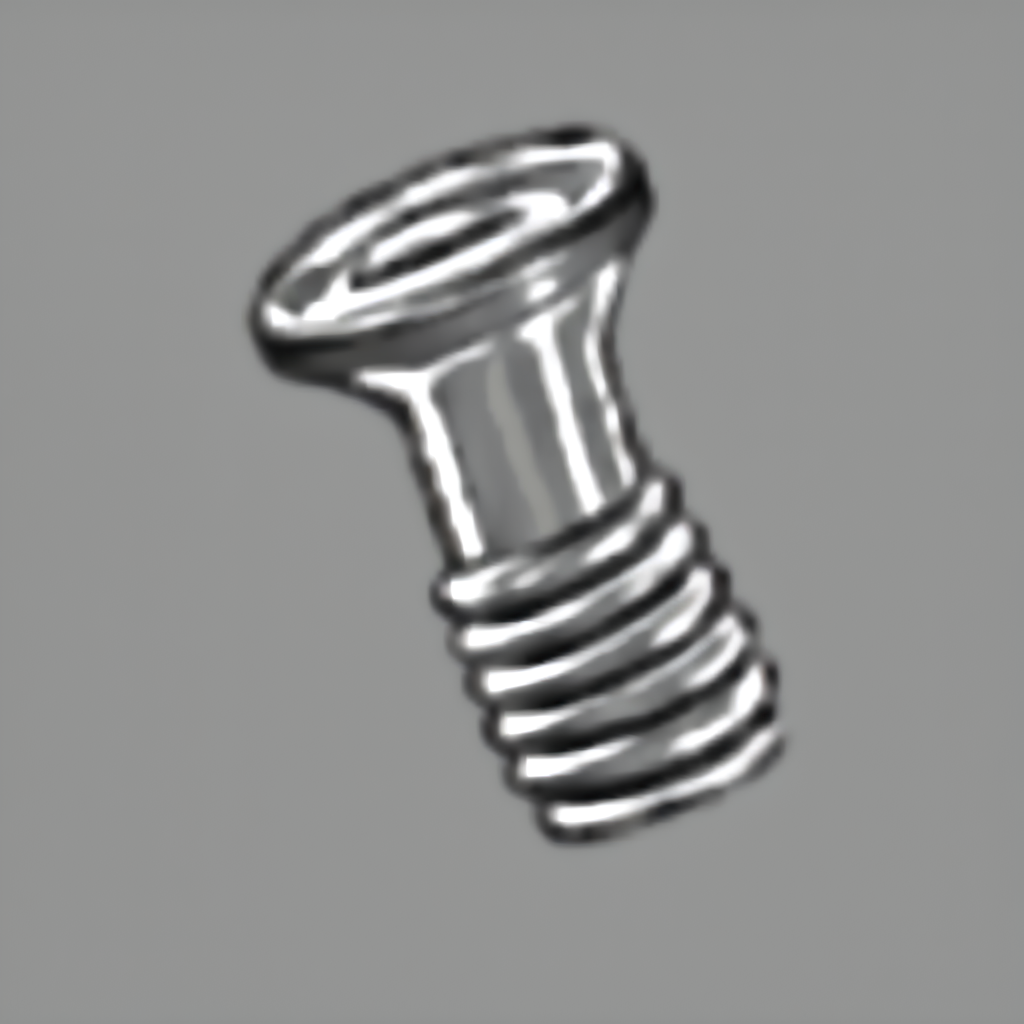}
        \caption{Spanner drive pan head screw icon}
        \label{fig:public_long_spanner_pan}
    \end{subfigure}
    \caption{Example generated icons based on public training data using long prompts}
    \label{fig:public_long_prompts}
\end{figure}

\FloatBarrier
\begin{figure}[H]
    \centering
    \begin{subfigure}{0.22\textwidth}
        \centering
        \includegraphics[width=\linewidth]{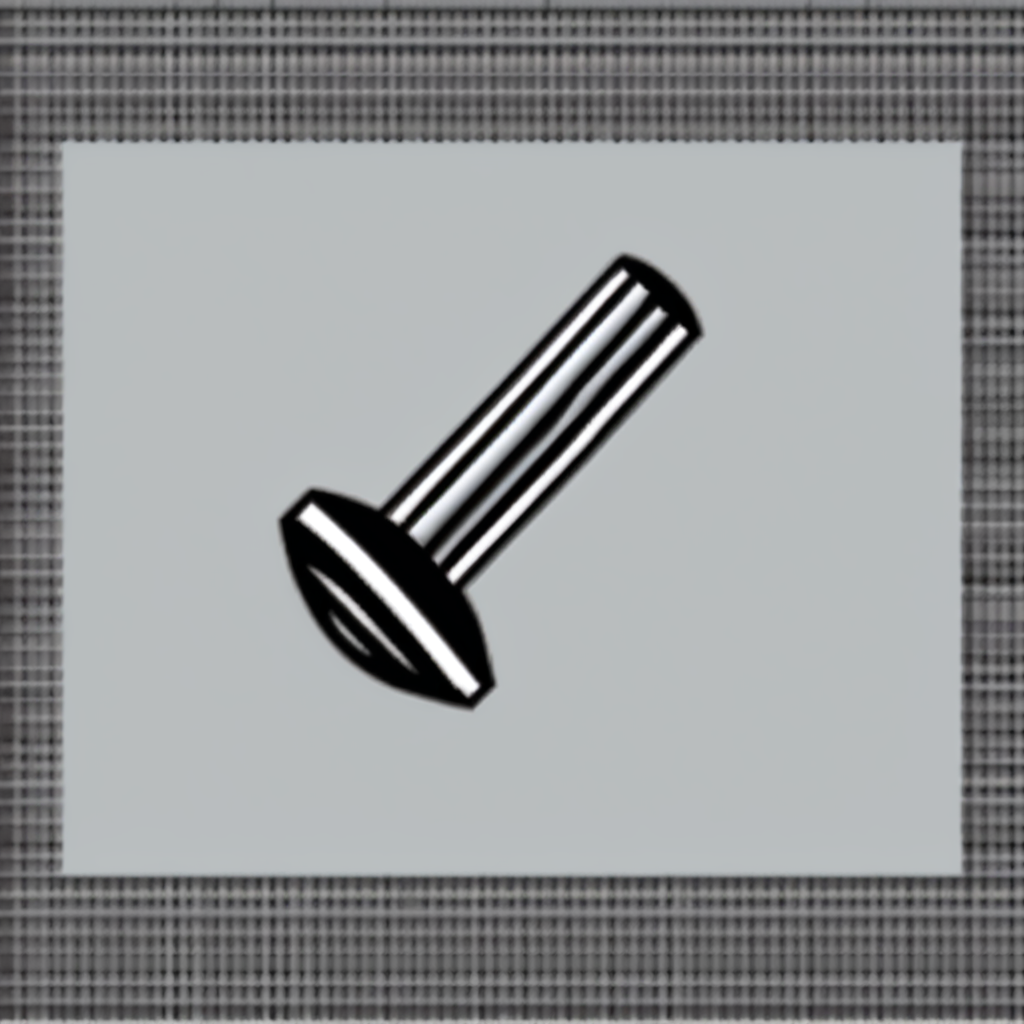}
        \caption{Hex head screw icon}
        \label{fig:public_hex_head}
    \end{subfigure}
    \hfill
    \begin{subfigure}{0.22\textwidth}
        \centering
        \includegraphics[width=\linewidth]{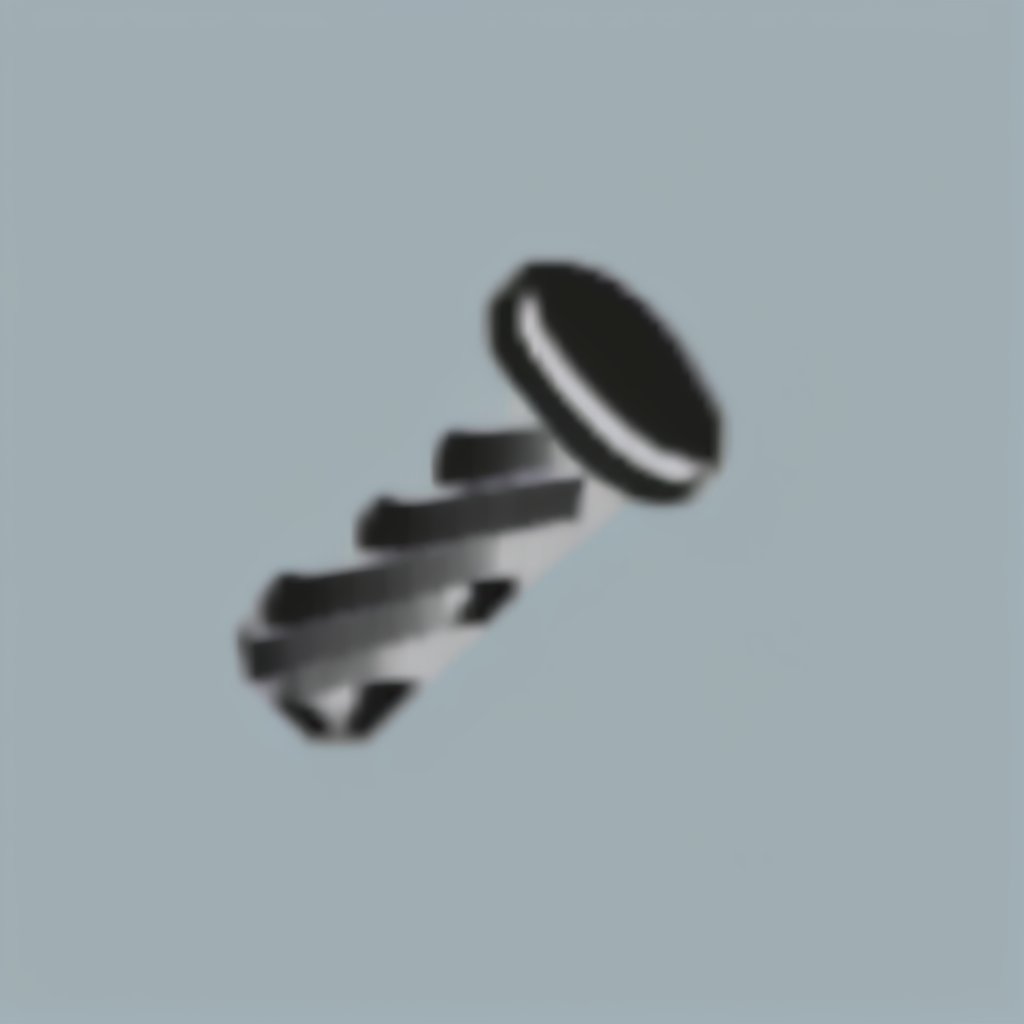}
        \caption{Round head screw icon}
        \label{fig:public_round_head}
    \end{subfigure}
    \hfill
    \begin{subfigure}{0.22\textwidth}
        \centering
        \includegraphics[width=\linewidth]{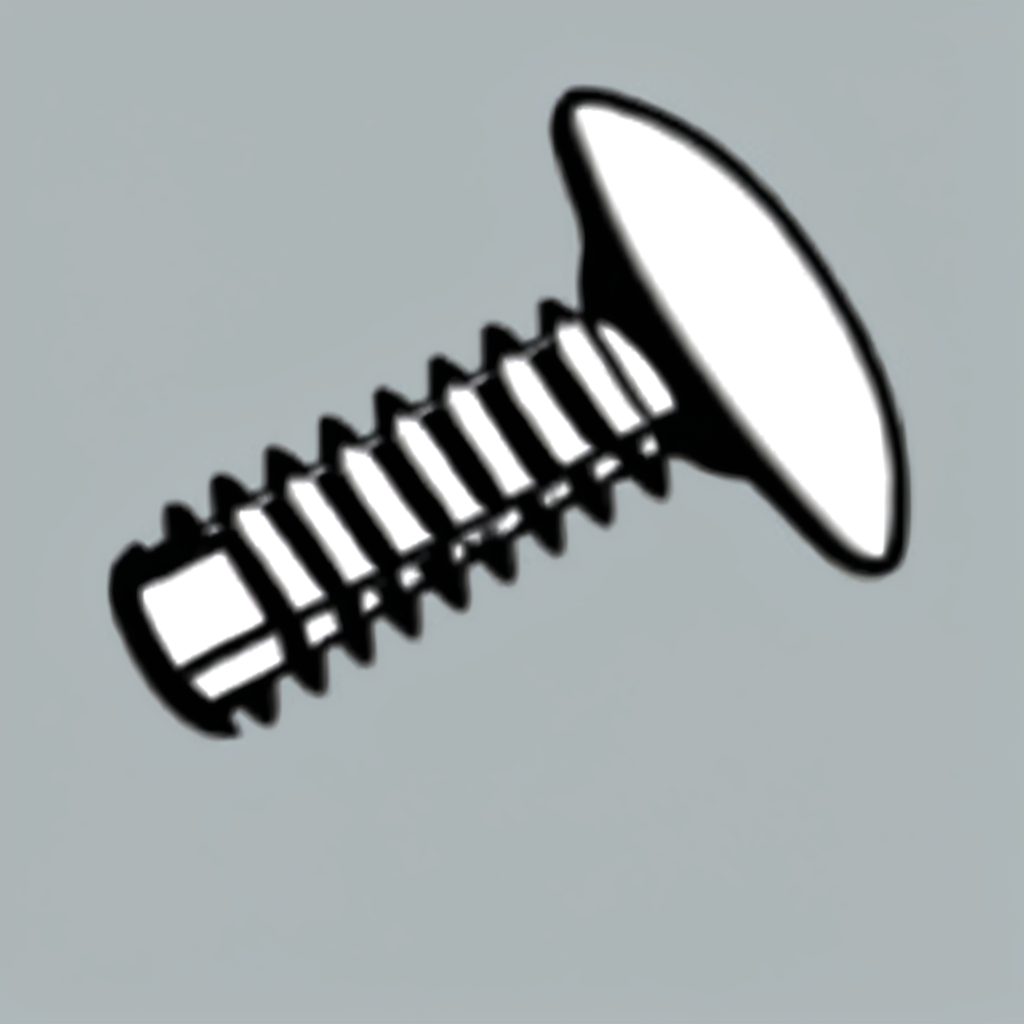}
        \caption{Phillips drive oval head screw icon}
        \label{fig:public_phillips_oval}
    \end{subfigure}
    \hfill
    \begin{subfigure}{0.22\textwidth}
        \centering
        \includegraphics[width=\linewidth]{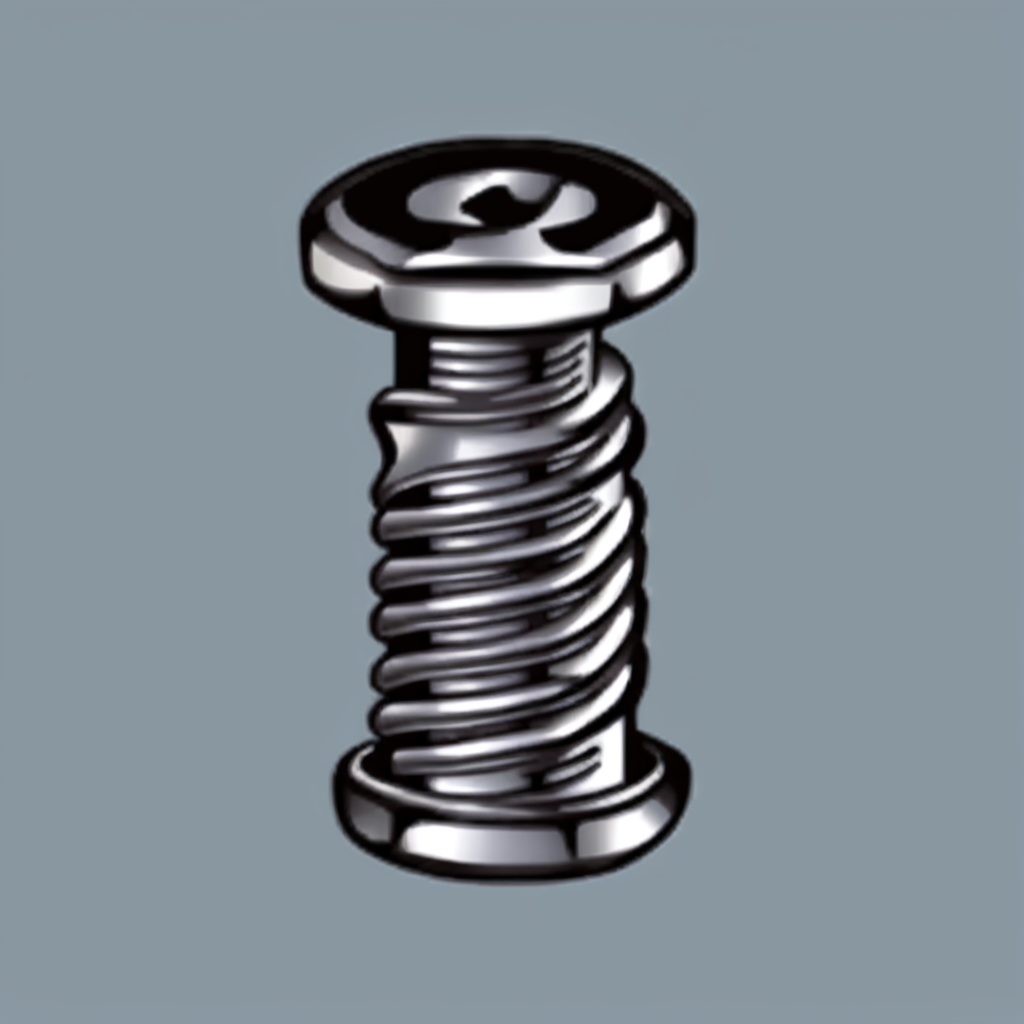}
        \caption{Spanner drive pan head screw icon}
        \label{fig:public_spanner_pan}
    \end{subfigure}
    \caption{Example generated icons based on public training data using short prompts}
    \label{fig:public_short_prompts}
\end{figure}

Based on the public dataset, the FID and CLIP scores were analyzed for both short and long prompts. The FID results show that the short prompt achieved a lower score of 227.69 compared to the long prompt's score of 263.48, indicating that the short prompt is better in terms of image quality and diversity. On the other hand, the CLIP scores were very close, with the short prompt scoring an average of 0.3214 and the long prompt scoring slightly higher at 0.3280. This suggests that while the short prompt performs better for FID, the long prompt has a marginally better alignment with the textual descriptions according to the CLIP score.

\FloatBarrier

\subsection{An analysis of inference steps and file-name keyword only training}

\begin{figure*}[htbp]
\centerline{\includegraphics[width=\textwidth]{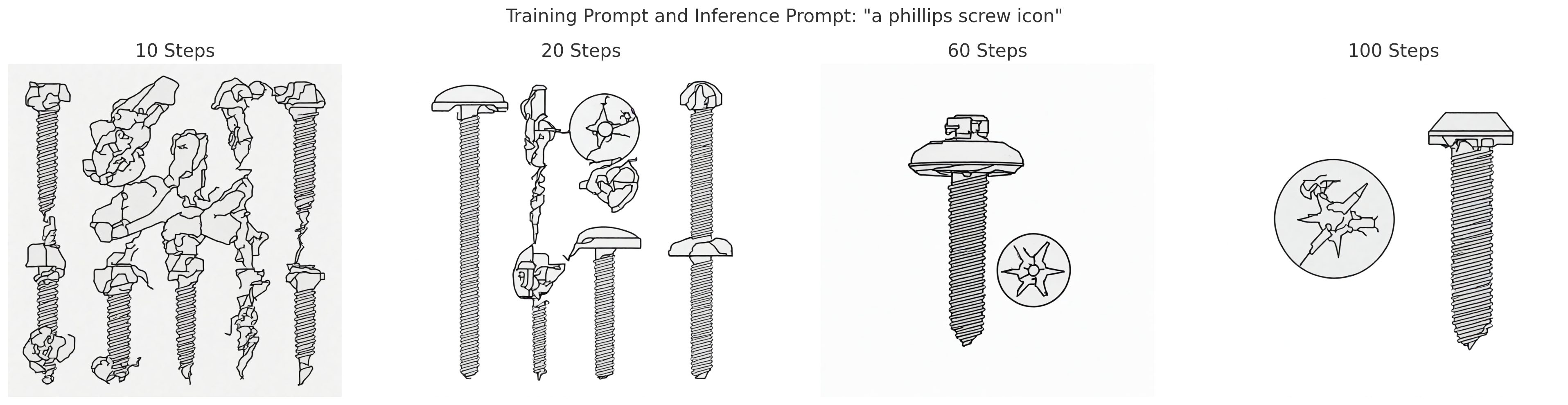}}
\caption{Testing different inference steps on keyword caption trained images}
\label{infsteps}
\end{figure*}

When testing the inference query on several different types of steps, it can be evident than an optimal point is required to find the best parameter value. As we increased the inference steps on a basic trained version of the model (sub-optimal) using keywords only for training, we found the image icons to get better as the inference steps increased. However there is a point where the inference steps being too high could cause a bad quality icon. The reasoning for the imperfections here ais due to smaller trained captions, smaller than our short captions with only file-name parsed keywords (see Figure~\ref{infsteps}).

\begin{figure}[htbp]
\centerline{\includegraphics[width=\linewidth]{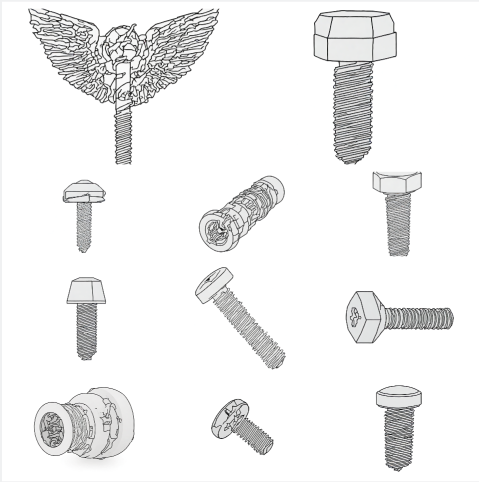}}
\caption{Various screw images trained on only relevant keywords from file names}
\label{infsteps3}
\end{figure}

When training the Stable Diffusion XL model on only keywords parsed from the file-names themselves, such as "Phillips head" or "hex screw", the results are not as good, however are still somewhat coherent. This tells us that without a unique identifier such as our use of "TOK", it is still possible to get the shape correctly for some cases, however some cases provide non-coherent icons (see Figure ~\ref{infsteps3}).
\FloatBarrier
\FloatBarrier

\FloatBarrier


\end{document}